\documentclass[10pt,twocolumn,letterpaper]{article}

\usepackage[pagenumbers]{cvpr} 

\usepackage[utf8]{inputenc} 
\usepackage[T1]{fontenc}    
\usepackage{url}            
\usepackage{booktabs}       
\usepackage{amsfonts}       
\usepackage{nicefrac}       
\usepackage{microtype}      
\usepackage[dvipsnames]{xcolor}         
\usepackage{natbib}       
\usepackage{tikz}
\usetikzlibrary{positioning, fit, fadings, 3d, spy}
\usetikzlibrary{shapes,arrows.meta,positioning,fit,backgrounds, calc}

\usepackage{amsmath}
\usepackage{siunitx}
\usepackage{lettrine}
\usepackage{calc}
\usepackage{wrapfig}
\usepackage{fontawesome5}
\usepackage{float}

\definecolor{cvprblue}{rgb}{0.21,0.49,0.74}
\usepackage[pagebackref,breaklinks,colorlinks,citecolor=cvprblue]{hyperref}

\def\paperID{*****} 
\def\confName{CVPR}
\def\confYear{2025}

\title{Geometry Image Diffusion: Fast and Data-Efficient Text-to-3D with Image-Based Surface Representation}

\author{%
  Slava Elizarov \\
  Unity Technologies\\
  {\tt\small slava.elizarov@unity3d.com}\\
  \and
  Ciara Rowles\\
  Unity Technologies\\
  {\tt\small ciara.rowles@unity3d.com}\\
  \and
  Simon Donn\'e\\
  Unity Technologies\\
  {\tt\small simon.donne@unity3d.com}
}

\begin{document}
\maketitle
\begin{abstract}
    Generating high-quality 3D objects from textual descriptions remains a challenging problem due to computational cost, the scarcity of 3D data, and complex 3D representations.
    We introduce \textbf{Geometry Image Diffusion} (GIMDiffusion), a novel Text-to-3D model that utilizes geometry images to efficiently represent 3D shapes using 2D images, thereby avoiding the need for complex 3D-aware architectures.
    By integrating a Collaborative Control mechanism, we exploit the rich 2D priors of existing Text-to-Image models such as Stable Diffusion.
    This enables strong generalization even with limited 3D training data (allowing us to use only high-quality training data) as well as retaining compatibility with guidance techniques such as IPAdapter.

    In short, GIMDiffusion enables the generation of 3D assets at speeds comparable to current Text-to-Image models.
    The generated objects consist of semantically meaningful, separate parts and include internal structures, enhancing both usability and versatility. 
\end{abstract}
\begin{figure*}
\centering
\input{figures/gim_generation/figure}
\caption{Meshes generated with our proposed Geometry Image Diffusion (GIMDiffusion) method. For each object, we show the generated albedo texture, the textured mesh, the untextured mesh, and the respective text prompt. The objects are generated entirely using our method: both the structure, texture and layout of the UV map are generated completely from scratch.}
\label{fig:gim-gen}
\end{figure*}

\section{Introduction}\label{sec:intro}%
Automatic 3D object generation promises significant benefits across video game production, cinema, manufacturing, and architecture. 
Despite notable progress in this area, particularly with Text-to-3D diffusion models~\cite{Boss2024SF3DSF, Siddiqui2024Meta3A, Wang2023ProlificDreamerHA}, generating high-quality 3D objects remains a challenging task due to computational costs, data scarcity, and the complexity of typical 3D representations. Furthermore, it is crucial that the generated objects can be re-lit within graphics pipelines, necessitating the use of physically-based rendering (PBR) materials.
Although graphics pipelines predominantly use meshes as their primary 3D representation, processing these at scale is notoriously difficult. Most techniques instead generate an intermediate representation, which adds additional burden to training data pre-processing and generated object post-processing.

We propose diffusing geometry images~\cite{Gu2002GeometryI}, a 2D representation of 3D surfaces, with a Collaborative Control scheme~\cite{Boss2024CollaborativeCF}.
This enables 3D object generation from text prompts, as shown in \cref{fig:gim-gen}.
The geometry image representation allows us to repurpose existing image-based architectures, while the Collaborative Control scheme enables us to leverage pre-trained Text-to-Image models, considerably reducing the required training data and cost.
Geometry images, and more specifically multi-chart geometry images, offer two great advantages over other representations: they do not impose constraints on the topology of the generated object, and they naturally separate the generated object into semantically meaningful parts, making the resulting objects easier to manipulate and edit.
We believe that GIMDiffusion opens up a promising new research direction in Text-to-3D generation, providing a practical and efficient approach that can inspire future advancements in the field.

In summary, the advantages of GIMDiffusion include:
\begin{itemize}
    \item \textbf{Image-based}: By leveraging existing 2D image-based models instead of developing new 3D architectures, we simplify both model design and training.
    \item \textbf{Fast Generation}: We generate well-defined 3D meshes in under 10 seconds per object, which could be further enhanced using distillation techniques.
    \item \textbf{Generalization}: Through collaborative control, we reuse pre-trained Text-to-Image priors, allowing us to generalize well beyond our limited 3D training data.
    \item \textbf{Separate Parts}: GIMDiffusion creates assets that consist of distinct, semantically-meaningful, separable parts, facilitating easier manipulation and editing of individual components.
    \item \textbf{Albedo Textures}: The 3D assets generated by GIMDiffusion do not have baked-in lighting effects, making them suitable for various environments.
    \item \textbf{Straightforward Post-processing}: Our 3D assets do not require the application of isosurface extraction algorithms or UV unwrapping, which reduces potential artifacts and simplifies the overall workflow.
\end{itemize}

\section{Related work}\label{sec:related}

\subsection{Text-to-Image Generation}
Diffusion models~\cite{Sohl-Dickstein2015-vg, Song2020ScoreBasedGM} and flow matching~\cite{Lipman2022-wr}, alongside the rise of versatile, general-purpose architectures such as transformers~\cite{Vaswani2017AttentionIA}, have brought considerable progress in generative modeling.
In particular, text-conditioned image generation was revolutionized by approaches based on Latent Diffusion~\cite{Rombach2021HighResolutionIS} and its further extensions~\cite{Podell2023SDXLIL, Pernias2024WrstchenAE, Esser2024ScalingRF}.
Foundational models like Stable Diffusion, trained on extensive internet-scale datasets (such as LAION-5B~\cite{Schuhmann2022LAION5BAO}), are capable of generating complex scenes from text prompts while exhibiting an implicit understanding of scene geometry.
Due to the high cost of training such models, they are often repurposed for other tasks or modalities~\cite{Zhang2023AddingCC, Hu2023AnimateAC, Boss2024CollaborativeCF}: our proposed GIMDiffusion is a prime example of this, restricting the base model to output albedo textures specifically.

\subsection{Conditioning Diffusion Models}
Control mechanisms modify pre-trained foundational models, enabling them to accept additional conditions to guide the generation process.
Existing pixel-aligned control techniques fall into two categories: fine-tuning the base model with modified input and output spaces~\cite{Duan2023DiffusionDepthDD, Ke2023RepurposingDI}, or a separate model that alters the base model's internal states~\cite{Zhang2023AddingCC, Zavadski2023ControlNetXSDA}.
The latter approaches, such as ControlNet~\cite{Zhang2023AddingCC}, have gained wide adoption due to their ability to preserve the original model's performance while adding conditions such as human poses or depth images.
AnimateAnyone~\cite{Hu2023AnimateAC} leverages a similar architecture to inject the base model's hidden states into a new branch that aligns to the base model's output.

In our application, we need to both control the base model (which will output UV-space albedo maps) and extract significant features from it (to generate the geometry image modality).
Collaborative Control~\cite{Boss2024CollaborativeCF} achieves exactly this by introducing bidirectional communication between both models, original for Text-to-PBR-Texture generation.
The ability to integrate completely new modalities into existing models, without retraining and with a reduced risk of catastrophic forgetting, is invaluable to our application. 

\subsection{Text-to-3D generation}

We identify two main approaches to Text-to-3D generation: optimization-based and feed-forward methods.
\emph{Optimization-based} methods rely on pre-trained 2D image diffusion models to generate 3D assets~\cite{Poole2022DreamFusionTU, Wang2023ProlificDreamerHA} and can produce content of high perceptual quality, at the cost of impractically long generation times~\cite{Lorraine2023ATT3DAT, Xie2024LATTE3DLA}.
These methods adapt existing 2D diffusion models to 3D by applying score distillation sampling~\cite{Poole2022DreamFusionTU, Wang2022ScoreJC} to iteratively optimize a 3D representation~\cite{Mildenhall2020NeRF,Kerbl20233DGS}. The key advantage of this approach is its ability to utilize the rich 2D prior, allowing for 3D object generation without the need for expensive 3D data.

However, the lack of camera conditioning leads to discrepancies among different viewing angles (Janus effect~\cite{Poole2022DreamFusionTU}) and projection artifacts, which require  3D-aware architectures and retraining on restrictive 3D datasets to mitigate these issues~\cite{Shi2023MVDreamMD, Liu2023Zero1to3ZO, Hllein2024ViewDiff3I, Zheng2023Free3DCN, Shi2023Zero123AS, Kant2024SPADS}, while weakening the 2D prior.
Additionally, the original formulation can lead to issues such as saturated colors, oversmooth geometry, and limited diversity~\cite{Wang2023ProlificDreamerHA, Zhu2023HIFAHT, Katzir2023NoiseFreeSD, Alldieck2024ScoreDS, Liang2023LucidDreamerTH, Wu2024Consistent3DTC}.

\emph{Feed-forward} methods directly generate 3D shapes without the need for iterative refinement.
However, we must consider compatibility with graphics pipelines, which predominantly use textured meshes as their primary representation — an inherently challenging modality for learning-based models.
While seminal works like Point-E~\cite{Nichol2022PointEAS} and its follow-ups~\cite{Huang2024PointInfinityRP,Zeng2022LIONLP} demonstrate impressive generalization and diversity, the inherent lack of connectivity information limits the expressiveness of point clouds. Instead, many current methods rely on proxy representations, such as neural implicits~\cite{Xie2021NeuralFI,Mildenhall2020NeRF,Malladi1995ShapeMW,Jun2023ShapEGC, Zheng2022SDFStyleGANIS, Chen2018LearningIF, Mescheder2018OccupancyNL} or neural triplanes~\cite{Hong2023LRMLR,Chan2021EfficientG3,Tochilkin2024TripoSRF3,Boss2024SF3DSF,bensadoun2024meta3dgen}, to represent the objects.

Because of this, these methods require pre- and post-processing to transform between the mesh domain and the proxy representation, \eg{} through marching cubes~\cite{Lorensen1987MarchingCA} or tetrahedra~\cite{Doi1991AnEM}. This process is costly and far from lossless, introducing issues such as quantization or grid-like artifacts, and resulting in the loss of information, including part segmentation and internal structures. While direct mesh generation techniques exist~\cite{siddiqui2023meshgpt}, these (as well as the other feed-forward method) must be trained from scratch on 3D data, which is scarce and expensive to gather.
By remaining in the image domain, GIMDiffusion  leverages Collaborative Control to ease this data burden~\cite{Boss2024CollaborativeCF}.

\subsection{Geometry Images}

Geometry images (GIM)~\cite{Gu2002GeometryI, Sander2003MultiChartGI} have been largely overlooked in deep learning~\cite{Sinha2016DeepL3}.
XDGAN~\cite{Alhaija2022XDGANM3} is a pioneering effort that utilizes GIMs as the representation of choice for a StyleGAN-based architecture~\cite{Karras2018ASG}.  However, due to the architectural constraints, the training data must be perfectly aligned with a template atlas,  which limits its applicability to real-world data. Furthermore, the pre-processing algorithm provided in the paper is overly restricted to shapes of genus zero --- \Cref{sec:preprocessing} shows how to handle arbitrary shapes.

Recent concurrent work~\cite{Yan2024AnOI} has advocated for low-resolution $64\times 64$ geometry images as a 3D representation for class-conditioned diffusion models and highlights its efficiency on a small-scale dataset of \num[group-separator={,}]{8000} objects~\cite{Collins2021ABODA}, albeit with limited generalization.
In contrast, GIMDiffusion tackles general Text-to-3D: rather than training a model from scratch, we leverage a pre-trained Text-to-Image diffusion model (using a Collaborative Control scheme trained on Objaverse~\cite{Deitke2022ObjaverseAU}) to retain generalization and diversity for the shapes, their appearance, \emph{and} the UV atlas layout.

\begin{figure*}[ht]
  \centering
  \input{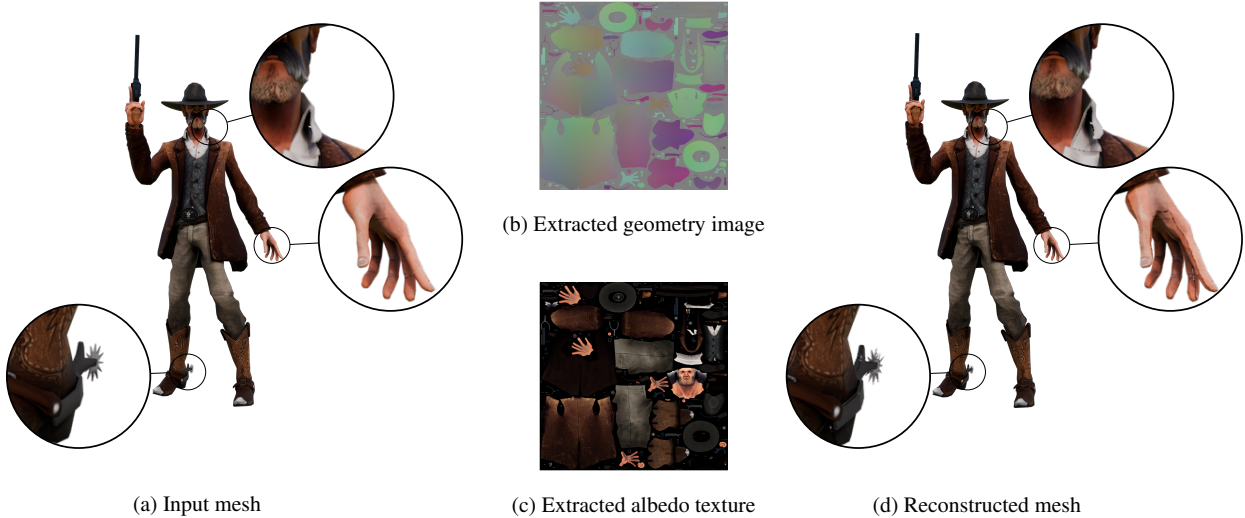}
  \caption{(a) Ground-truth geometry, (b) geometry image and (c) albedo texture from our data pre-processing, and (d) the reconstruction using our dedicated VAE. We note the highly separable nature of the ground truth object, which is split into small components. The only visible artifact after decoding is the missing connection between the various charts of the geometry image, as discussed in \cref{sec:data-handling,sec:limitations}.}\label{fig:gim_rec}
\end{figure*}

\section{Method}\label{sec:method}

\subsection{Geometry Images}\label{sec:gim}

Geometry images~\cite{Gu2002GeometryI} represent a 3D surface in image format as a function $\phi : [0, 1]^2 \rightarrow S \in \mathbb{R}^3$, where $S$ is the object surface and $\left[0, 1\right]^2$ represents the UV coordinates in the unit square, typically sampled on a uniform grid of the desired resolution.
The choice of the function $\phi$ is critical and is usually designed to minimize spatial distortion (using conformal mapping) for a uniform covering of the geometry.

Unlike a traditional mesh, which requires explicit data structures to maintain connectivity information to form faces, geometry images implicitly connect each pixel to its immediate neighbors.
While Gu \etal{} cut the input surface and warp it onto a disc~\cite{Gu2002GeometryI} (\ie{} restricted to manifold objects,  resulting in significant distortions for high-genus objects), Multi-Chart Geometry Images~\cite{Sander2003MultiChartGI} address this limitation by using atlas construction to map the surface piecewise onto several charts of arbitrary shape (each chart being homeomorphic to a disc). This approach adds flexibility (removing the manifold constraint), and reduces distortion (providing greater geometric fidelity), though the optimal construction of such multi-chart GIMs remains an active area of research~\cite{Sawhney2017BoundaryFF, Srinivasan2023NuvoNU}. 

Fortunately, most available 3D meshes have textures applied through UV maps. We observe that these handcrafted UV maps can be employed to construct a desirable multi-chart $\phi$.
Furthermore, charts in handcrafted UV maps often carry semantic meanings, which propagate to the output of our model. This is evident in \cref{fig:gim_rec}, where the hands, face, and various part of the gunslinger's appearance are clearly separated in the UV atlas.

As the density of the generated geometry's triangulation is restricted by the resolution of the underlying geometry image, we follow~\cite{Rombach2021HighResolutionIS} and use a VAE to increase the effective resolution of our proposed model.
To address the irregularities in geometry images and better match their distribution — particularly the need to accurately reconstruct the discontinuities at the boundaries of the charts — we add a channel to represent the multi-chart mask and modify the loss function accordingly.
Otherwise, we follow the VAE training procedure from StableDiffusion1.5~\cite{Rombach2021HighResolutionIS}.
As shown in \cref{fig:gim_rec}, the reconstruction using this VAE does not visually differ from the input mesh, except for the missing connections between the charts, which we do not currently handle.

\subsection{Collaborative control}\label{sec:collaborative_control}

\begin{figure*}
  \input{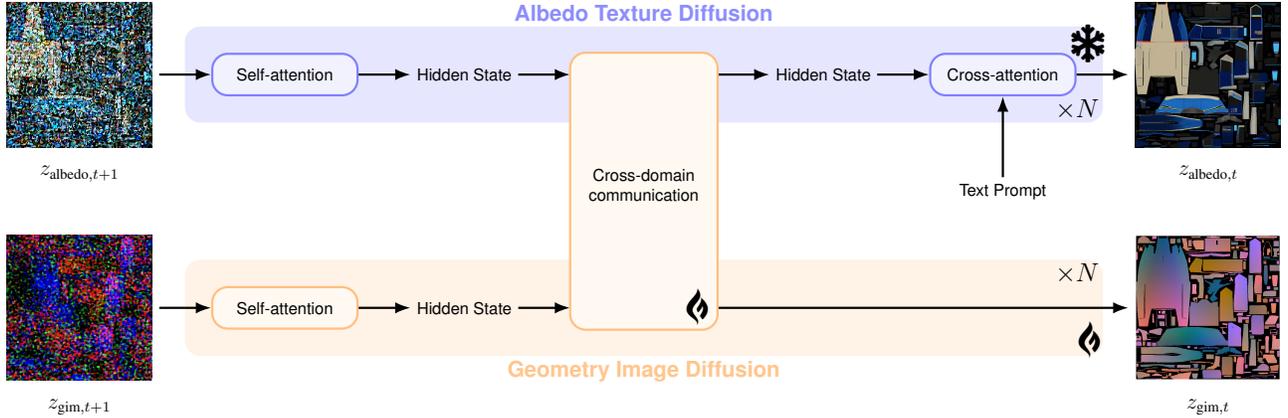}
  \caption{The Collaborative Control Scheme~\cite{Boss2024CollaborativeCF} applied in GIMDiffusion, where two separate diffusion models generate respectively albedo textures and geometry images. The former is a frozen pre-trained model, while the latter is an architectural clone trained from scratch.}\label{fig:collab_control}
\end{figure*}

To maximally leverage the prior knowledge encoded in existing 2D Text-to-Image models, we use the Collaborative Control approach~\cite{Boss2024CollaborativeCF}. As shown in \cref{fig:collab_control}, this approach comprises two parallel networks: a pre-trained RGB model and a new model for the geometry image. The former is responsible for generating UV-space albedo textures, while the latter generates the geometry images. These two models are connected by a simple linear cross-network communication layer, which allows them to share information and collaborate in generating pixel-aligned outputs across these different modalities. Crucially, this also enables the geometry model to influence the frozen model,  guiding it to generate UV-space textures that would otherwise lie at the fringes of its training distribution. The frozen base model also drastically reduces the amount of data required to train the joint model while retaining generalizability, diversity, and quality~\cite{Boss2024CollaborativeCF}.

\subsection{Data Handling}\label{sec:data-handling}

\subsubsection{Dataset}\label{sec:dataset}

We train our model on the Objaverse dataset~\cite{Deitke2022ObjaverseAU}.
We curate this dataset to include only objects with both high-quality structures and semantically meaningful UV maps by filtering out 3D scans and low-poly models.
The final dataset contains approximately \num[group-separator={,}]{100000} objects.
Each data entry is accompanied by captions provided by Cap3D~\cite{luo2023scalable} and Hong \etal{}~\cite{Hong20243DTopiaLT}.
During training, we randomly sample from these captions and apply random rotations of 90, 180, or 270 degrees to the extracted texture atlases.
We now discuss how to transform these meshes into geometry images and back: the entire preprocessing was performed on consumer-grade PC hardware (AMD Ryzen 9 7950X, GeForce RTX 3090, 64 GB RAM) and took approximately 20 hours.

\subsubsection{Geometry Image Creation}\label{sec:preprocessing}

\begin{figure}
  \begin{center}
      \begin{minipage}[c]{0.5\linewidth}
        \resizebox{\linewidth}{!}{
\begin{tikzpicture}[
    x={(0.866cm,-0.5cm)}, 
    y={(0.866cm,0.5cm)}, 
    z={(0cm,1cm)},
    vertex/.style={circle,fill=AccentBlue,inner sep=1.5pt},
    edge/.style={thick,draw=AccentBlue},
    label/.style={font=\small\sffamily}
]

\definecolor{PrimaryBlue}{HTML}{B0D4E5}
\definecolor{AccentBlue}{HTML}{8ABCD1}
\definecolor{NeutralBackground}{HTML}{F5F5F5}
\definecolor{PrimaryGreen}{HTML}{A5D6A7}
\definecolor{AccentOrange}{HTML}{FFB74D}


\coordinate (A) at (0, 0, {0.5*sqrt(1-((0-2)^2+(0)^2)/16)});
\coordinate (B) at (2, 0, {0.5*sqrt(1-((2-2)^2+(0)^2)/16)});
\coordinate (C) at (1, 1.5, {0.5*sqrt(1-((1-2)^2+(1.5)^2)/16)});
\coordinate (D) at (3, 1.5, {0.5*sqrt(1-((3-2)^2+(1.5)^2)/16)});
\coordinate (E) at (1, -1.5, {0.5*sqrt(1-((1-2)^2+(-1.5)^2)/16)});
\coordinate (F) at (4, 0, {0.5*sqrt(1-((4-2)^2+(0)^2)/16)});
\coordinate (G) at (3, -1.5, {0.5*sqrt(1-((3-2)^2+(-1.5)^2)/16)});

\foreach \triangle in {
    {(A) -- (B) -- (C)},
    {(B) -- (D) -- (C)},
    {(A) -- (B) -- (E)},
}{
    \fill[AccentOrange!70,opacity=0.8] \triangle -- cycle;
    \draw[edge] \triangle -- cycle;
}

\foreach \triangle in {
    {(B) -- (F) -- (D)},
    {(E) -- (G) -- (B)},
    {(B) -- (G) -- (F)}
}{
    \fill[PrimaryGreen!70,opacity=0.8] \triangle -- cycle;
    \draw[edge] \triangle -- cycle;
}

\draw[red, thick, dashed] (E) -- (B) -- (D);

\node[anchor=north,inner sep=5pt,rotate=270] at (D.north) {\huge\faCut};

\foreach \point/\pos in {A/below,B/below,C/above,D/above,E/below,F/right,G/below right}{
    \if\point E
        \node[vertex, red] at (\point) {};
    \else
        \if\point B 
            \node[vertex, red] at (\point) {};
        \else
                \if\point D
                    \node[vertex, red] at (\point) {};
                \else
                    \node[vertex] at (\point) {};
                \fi
        \fi
    \fi
}

\end{tikzpicture}
}
      \end{minipage}
      \hfill
      \begin{minipage}[c]{0.48\linewidth}
        \resizebox{\linewidth}{!}{
\begin{tikzpicture}[
    x={(0.866cm,-0.5cm)}, 
    y={(0.866cm,0.5cm)}, 
    z={(0cm,1cm)},
    vertex/.style={circle,fill=AccentBlue,inner sep=1.5pt},
    edge/.style={thick,draw=AccentBlue},
    label/.style={font=\small\sffamily}
]

\definecolor{PrimaryBlue}{HTML}{B0D4E5}
\definecolor{AccentBlue}{HTML}{8ABCD1}
\definecolor{NeutralBackground}{HTML}{F5F5F5}


\coordinate (A) at (0, 0, {0.5*sqrt(1-((0-2)^2+(0)^2)/16)});
\coordinate (B) at (2, 0, {0.5*sqrt(1-((2-2)^2+(0)^2)/16)});
\coordinate (C) at (1, 1.5, {0.5*sqrt(1-((1-2)^2+(1.5)^2)/16)});
\coordinate (D) at (3, 1.5, {0.5*sqrt(1-((3-2)^2+(1.5)^2)/16)});
\coordinate (E) at (1, -1.5, {0.5*sqrt(1-((1-2)^2+(-1.5)^2)/16)});
\coordinate (F) at (4, 0, {0.5*sqrt(1-((4-2)^2+(0)^2)/16)});
\coordinate (G) at (3, -1.5, {0.5*sqrt(1-((3-2)^2+(-1.5)^2)/16)});

\foreach \triangle in {
    {(A) -- (B) -- (C)},
    {(B) -- (D) -- (C)},
    {(A) -- (B) -- (E)},
    {(B) -- (F) -- (D)},
    {(E) -- (G) -- (B)},
    {(B) -- (G) -- (F)}
}{
    \fill[PrimaryBlue!70,opacity=0.8] \triangle -- cycle;
    \draw[edge] \triangle -- cycle;
}

\draw[red, thick, dashed] (E) -- (B) -- (D);

\node[anchor=north,inner sep=5pt,rotate=270] at (D.north) {\huge\faCut};

\foreach \point/\pos in {A/below,B/below,C/above,D/above,E/below,F/right,G/below right}{
    \if\point E
        \node[vertex, red] at (\point) {};
        \node[label,\pos=2pt] at (\point) {1};
    \else
        \if\point B 
            \node[vertex, red] at (\point) {};
            \node[label,\pos=2pt] at (\point) {1};
        \else
                \if\point D
                    \node[vertex, red] at (\point) {};
                    \node[label,\pos=2pt] at (\point) {1};
                \else
                    \node[vertex] at (\point) {};
                    \node[label,\pos=2pt] at (\point) {2};
                \fi
        \fi
    \fi
}

\end{tikzpicture}
}
      \end{minipage}
      \caption{Seam detection in our multi-chart geometry image creation procedure to isolate \emph{locally invertible} areas of the UV mapping. (Left) If two neighboring mesh regions correspond to two distinct charts in the UV map, the vertices on the boundary will be duplicated and have different UV coordinates. (Right) If the UV mapping loops back onto itself, there will be a local minimum in the UV access heatmap, and we place the seam along the line with the smallest UV-degree to effectively separate these regions.}\label{fig:uv_degree}
  \end{center}
  \end{figure}
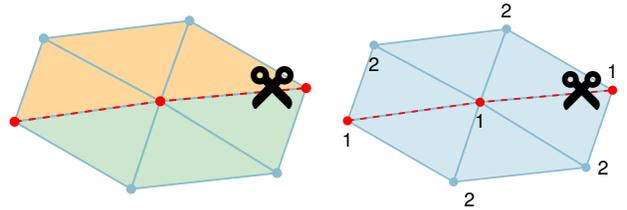

To create multi-chart geometry images for the training shapes, we use their existing UV maps.
UV mapping is defined as the mapping $\rho: V \rightarrow \left[0, 1\right]^2$, where $V$ is the set of vertices in a 3D mesh and $\rho$ maps each vertex to UV coordinates.
Note that this mapping is not injective (multiple vertices can be mapped to the same UV coordinates), nor is it a simple function of the vertex positions in $\mathbb{R}^3$, as modern mesh formats allow for multiple vertices with distinct UV coordinates at the same 3D location (to handle sharp texture transitions).
These issues mean that $\rho$ is not invertible, which would be a simple way to create a geometry image function.
However, we argue that $\rho$ is \emph{locally invertible} and propose to construct a multi-chart geometry image based on the individually invertible areas of the available UV mapping.
As mentioned before, the islands in a UV atlas tend to be semantically meaningful, so that we aim to preserve them.

We begin by identifying the connected components of the mesh, which provides an initial separation into charts. Within each component, we identify two situations where $\rho$ is not invertible: duplicated vertices with distinct UV coordinates, and a ``crease'', \ie{} a line where the UV coordinates change direction (similar to a ``mirror'' boundary constraint). Both situations are visualized in \cref{fig:uv_degree}. The former is straightforward to detect, as we can detect duplicated vertex positions. The latter is detected by creating a heatmap of the access pattern in UV space and detecting local minima. We then further split the individual charts along any detected creases.
Finally, in line with the desirable mapping properties discussed in~\cite{Sinha2016DeepL3}, we adjust the geometry image mapping to approximate an equal-area projection by rescaling each 2D chart with respect to the area of the corresponding surface.

In cases where only a partial UV mapping is available, we use Xatlas~\cite{Young2022} to UV-unwrap the missing regions. However, since the Xatlas parametrization is of lower quality and lacks semantic properties, we exclude meshes where less than $80\%$ of the surface area has been unwrapped manually. This simple heuristic allows us to construct multi-chart geometry images for nearly all training examples.
However, it does not account for all possible degenerate cases of $\rho$.
Therefore, we verify that the constructed $\rho$ is injective and skip any training meshes for which this assumption is violated.

\subsubsection{Coordinate transform}

In Objaverse~\cite{Deitke2022ObjaverseAU}, the objects are not aligned in a canonical way.
Although almost all shapes are oriented with the Y axis pointing upward, there is no fixed front view.
In practice, this leads to a rotational ambiguity between the X and Z axes.
To resolve this ambiguity and isolate it to a single coordinate, we leverage cylindrical coordinates $\left(r,\theta,\phi\right)$ so that the ambiguity is contained within the azimuth $\theta$.

\begin{figure}
  \centering
  
  \resizebox{\linewidth}{!}{
  \begin{tikzpicture}
  [spy using outlines={circle, magnification=5, size=2cm, connect spies}]

    \node[inner sep=0pt] (image) {\includegraphics[width=6cm]{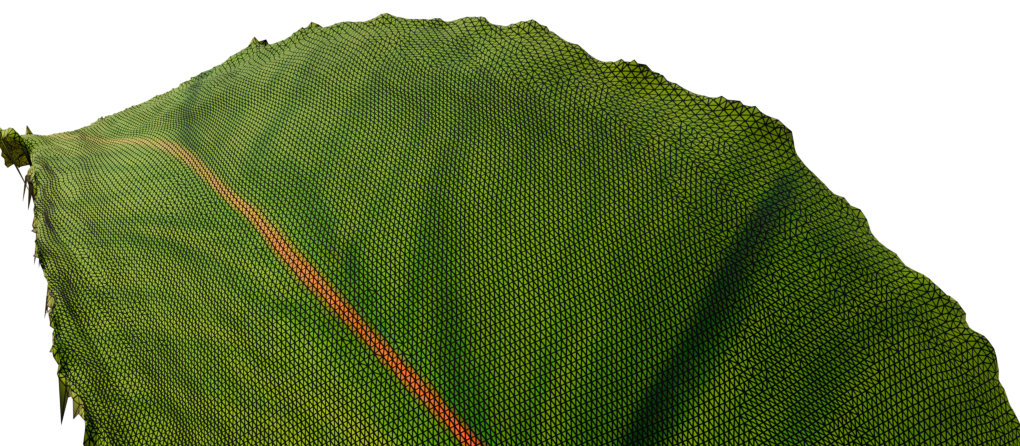}};

    \spy on ($(image) + (-1.3,0)$) in node at ($(image) + (2,0)$);
  \end{tikzpicture}
  }
  \caption{The resulting triangulation of our generated objects is near-uniform over the surface, thanks to the area-preserving nature of the geometry images in our training dataset.}\label{fig:wireframe}
\end{figure}

\subsubsection{Mesh Extraction}\label{sec:triangulation}

As mentioned in \cref{sec:gim}, geometry images implicitly encode connectivity information of the mesh by treating neighboring pixels as connected in 3D space, \ie{} a quad mesh.
However, to convert them into a more widely supported triangular mesh, we need to specify exactly how to triangulate the quads.
For this, we closely follow the algorithm from Sander\etal{}~\cite{Sander2003MultiChartGI}.
For any $2\times 2$ block of pixels in the GIM, we create up to two triangles depending on how many of the pixels are valid: if necessary, the quad is split along its shorter diagonal.
As shown in \cref{fig:wireframe}, and in line with our goal of area-preserving mapping~\cite{Sinha2016DeepL3}, the resulting triangulation is nearly uniform over the surface, which may or may not be desirable for specific applications (as the working resolution of our model is $768\times 768$, our GIMs can encode meshes with up to \num[group-separator={,}]{589824} vertices).
We consider the generation of arbitrary topologies or with a polygon constraint a promising area of future work for GIMDiffusion. 

\subsection{Training}\label{sec:training}
For the frozen base model, we used a zero-terminal-SNR~\cite{lin2024zerosnr} version fine-tuned~\cite{drhead2024ZeroDiffusion} from StableDiffusion v2.1~\cite{Rombach2021HighResolutionIS} as the base Text-to-Image model, which is kept frozen and generates the albedo textures. The geometry model is an architectural clone that is trained from scratch, together with the cross-network communication layers. Initially, we train the model at $256\times 256$ resolution for \num[group-separator={,}]{250000} steps with a batch size of $48$, and then at the final output resolution of $768\times 768$ for a total of \num[group-separator={,}]{100000}  steps with a batch size of $8$.
All stages of training were conducted with a learning rate of $3e^{-5}$ on 8 A100 GPUs.

\section{Results}

\begin{figure*}\centering
\begin{minipage}[t]{0.032\linewidth}
    \centering
    \rotatebox{90}{\parbox{4.8\linewidth}{\centering Albedo Texture}}

    \rotatebox{90}{\parbox{4.8\linewidth}{\centering Textured Mesh}}

    \rotatebox{90}{\parbox{4.8\linewidth}{\centering Untextured Mesh}}
\end{minipage}\hfill
\begin{minipage}[t]{0.48\linewidth}
\begin{minipage}[t]{0.32\linewidth}
    \centering
    \includegraphics[width=\linewidth,height=\linewidth]{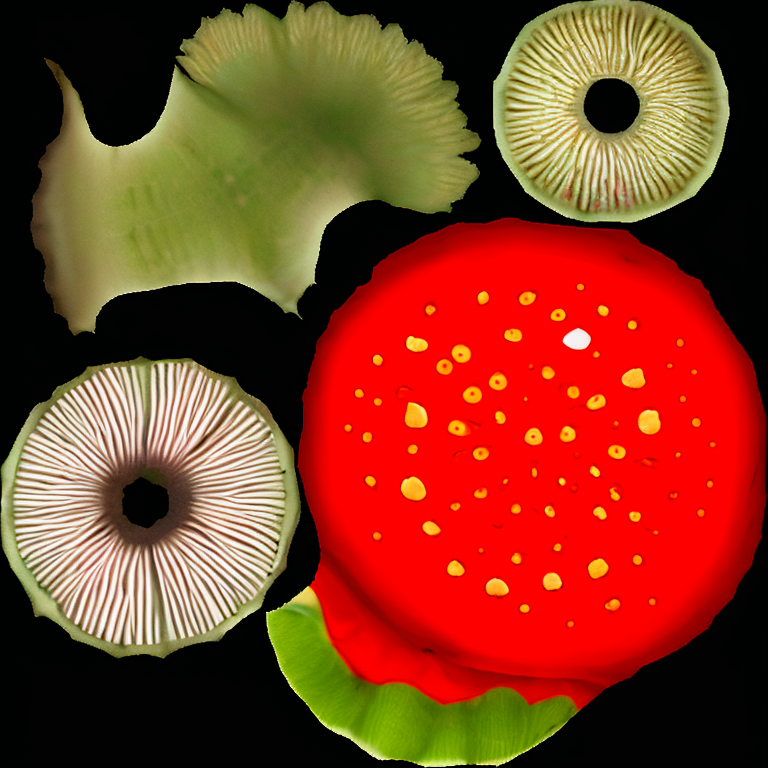}
    \includegraphics[width=\linewidth,height=\linewidth]{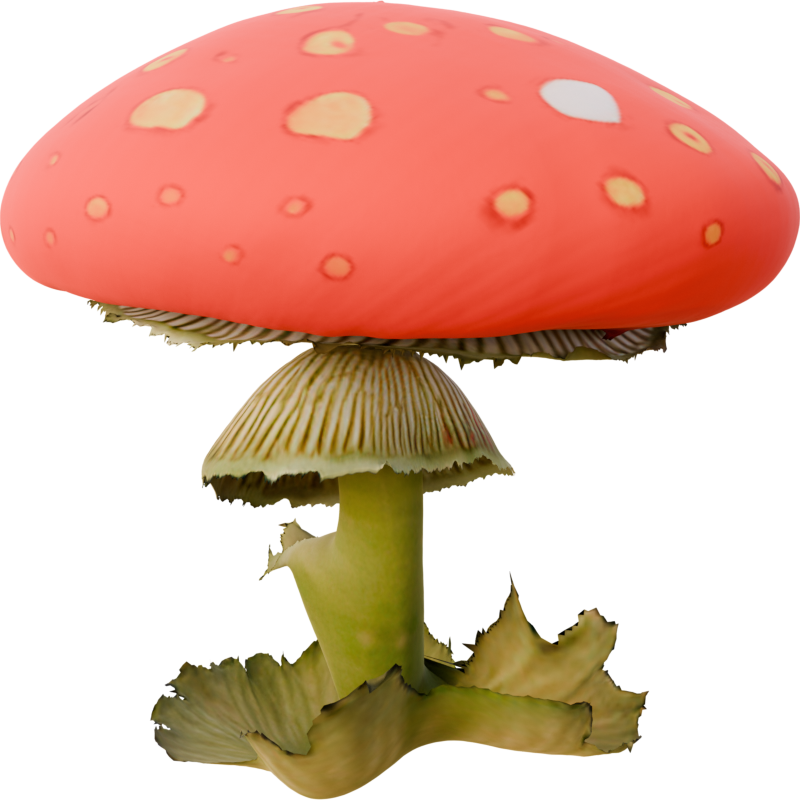}
    \includegraphics[width=\linewidth,height=\linewidth]{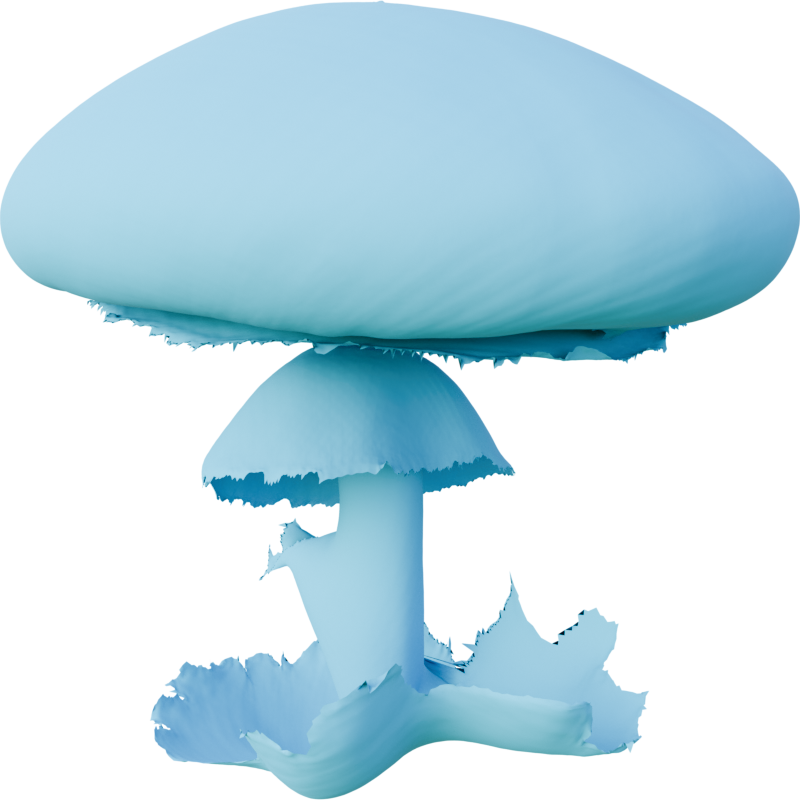}\vspace{1mm}
    \parbox{\linewidth}{ \centering\small\textbf{Seed: } \textit{906412}}
\end{minipage}\hfill
\begin{minipage}[t]{0.32\linewidth}
    \centering
    \includegraphics[width=\linewidth,height=\linewidth]{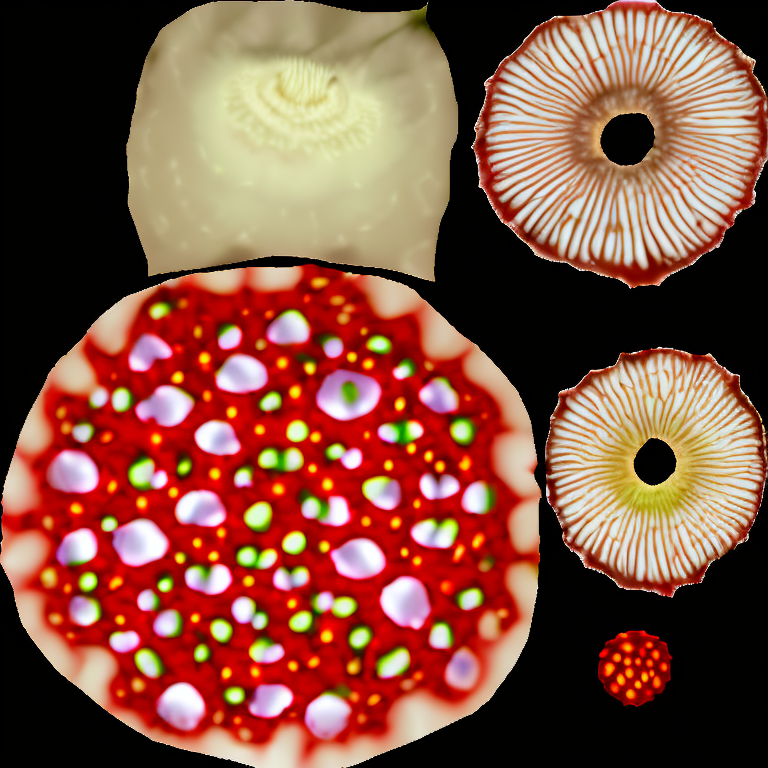}
    \includegraphics[width=\linewidth,height=\linewidth]{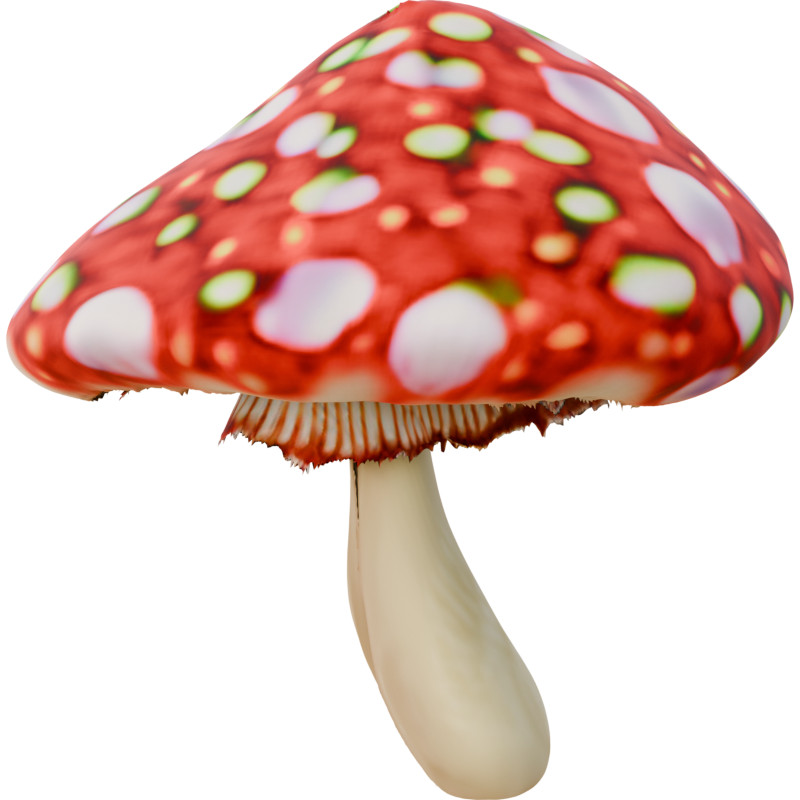}
    \includegraphics[width=\linewidth,height=\linewidth]{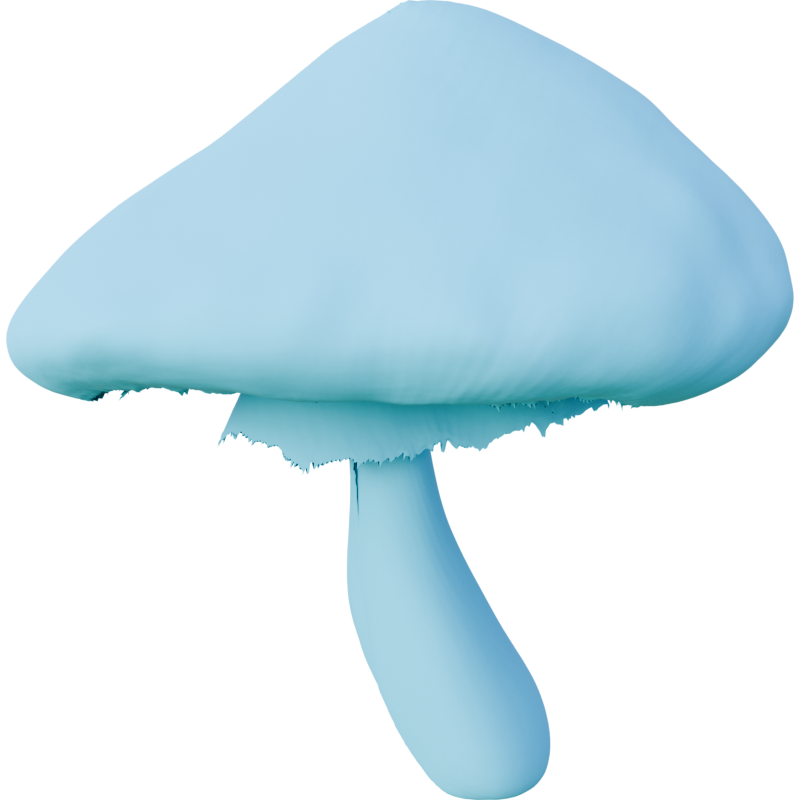}\vspace{1mm}
    \parbox{\linewidth}{ \centering\small\textbf{Seed:} \textit{745785}}
\end{minipage}\hfill
\begin{minipage}[t]{0.32\linewidth}
    \centering
    \includegraphics[width=\linewidth,height=\linewidth]{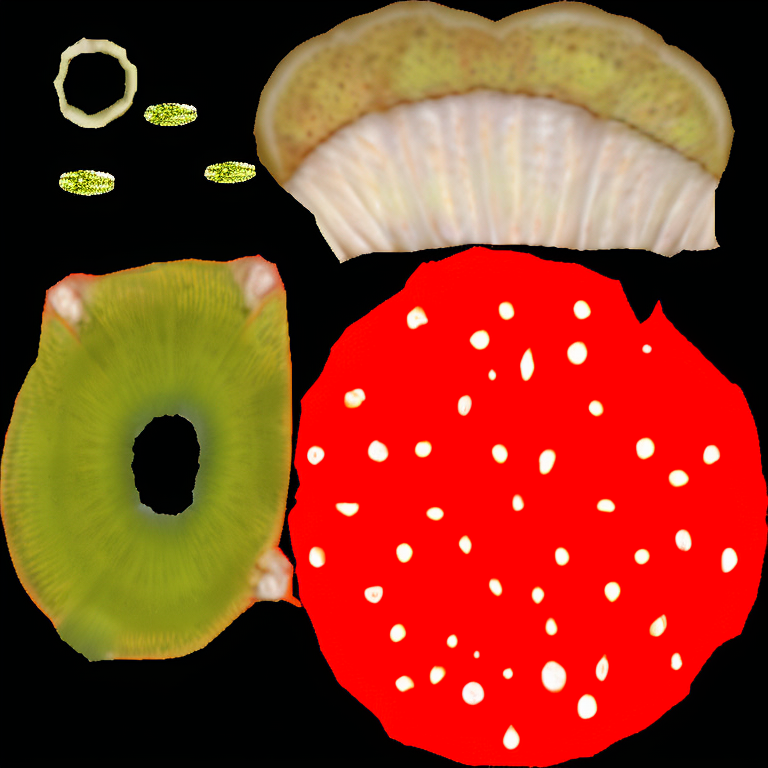}
    \includegraphics[width=\linewidth,height=\linewidth]{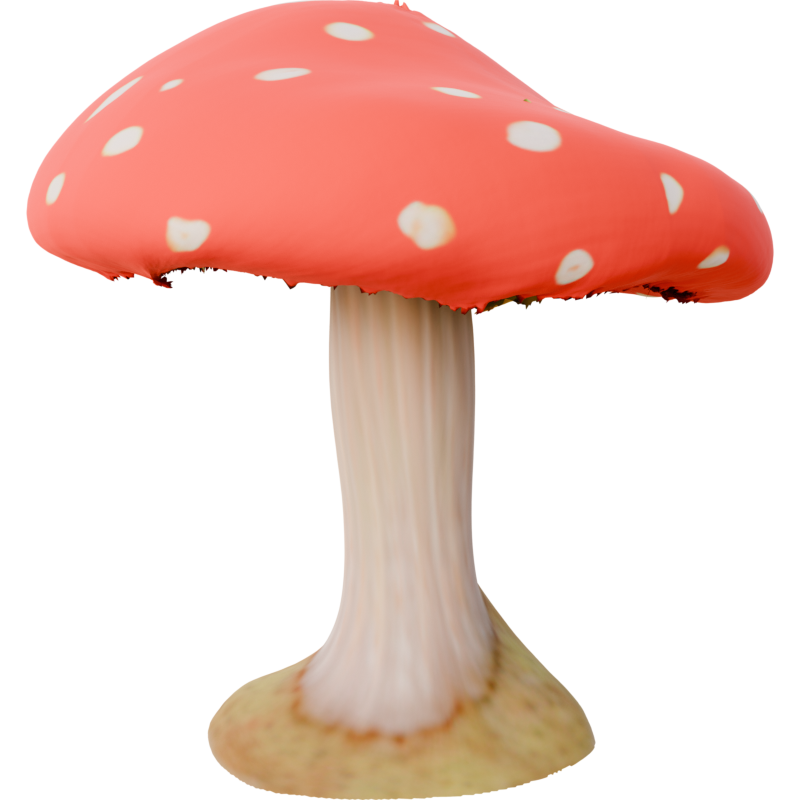}
    \includegraphics[width=\linewidth,height=\linewidth]{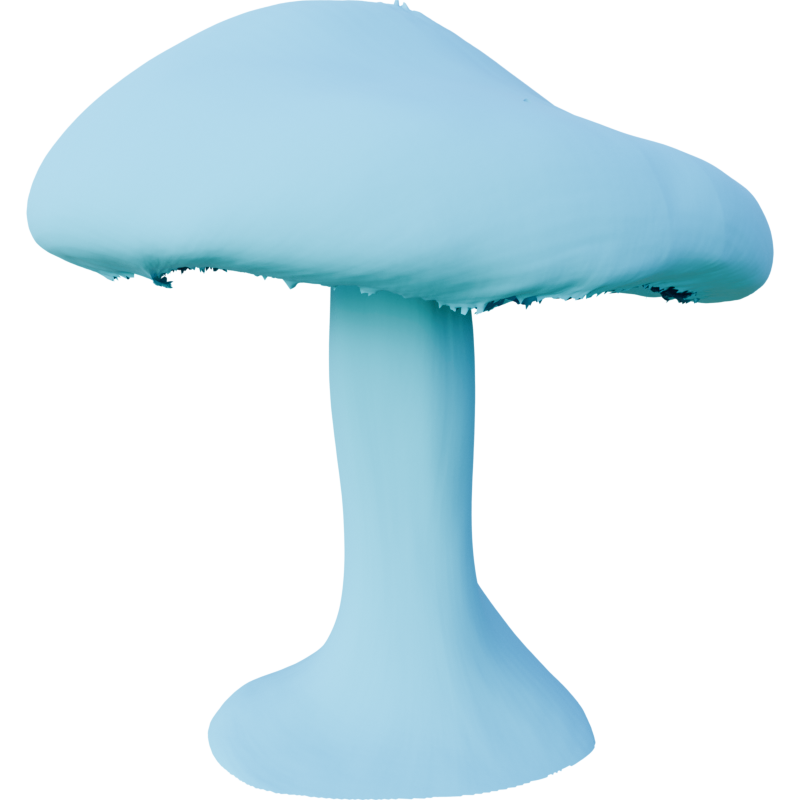}\vspace{1mm}
    \parbox{\linewidth}{ \centering\small\textbf{Seed:} \textit{149311}}
\end{minipage}\vspace{2mm}
\begin{minipage}{0.9\linewidth}
    \centering\small\textbf{Prompt:} \textit{A fly agaric mushroom}
\end{minipage}
\end{minipage}
\hfill
\begin{minipage}[t]{0.48\linewidth}
\begin{minipage}[t]{0.32\linewidth}
    \centering
    \includegraphics[width=\linewidth,height=\linewidth]{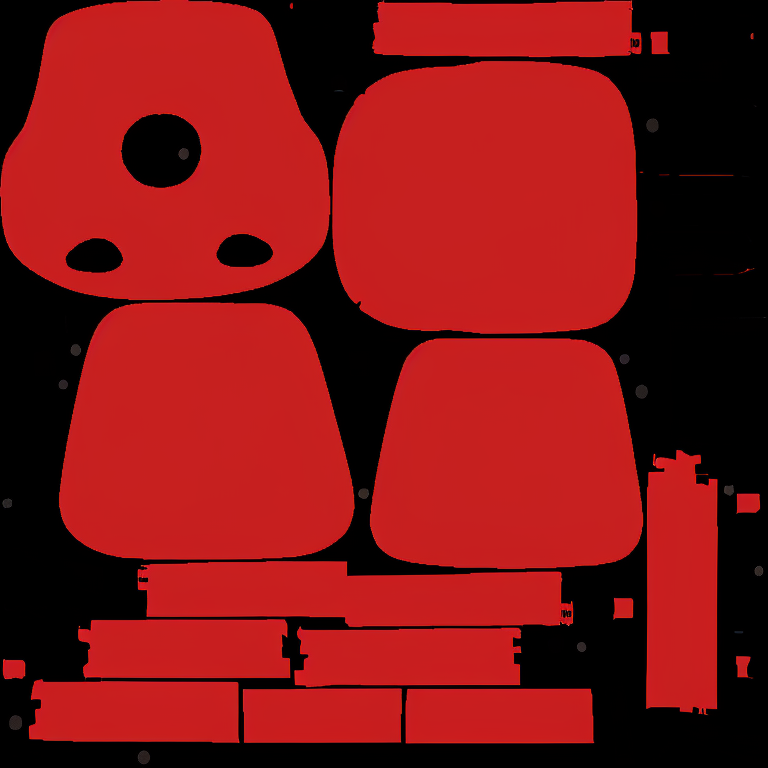}
    \includegraphics[width=\linewidth,height=\linewidth]{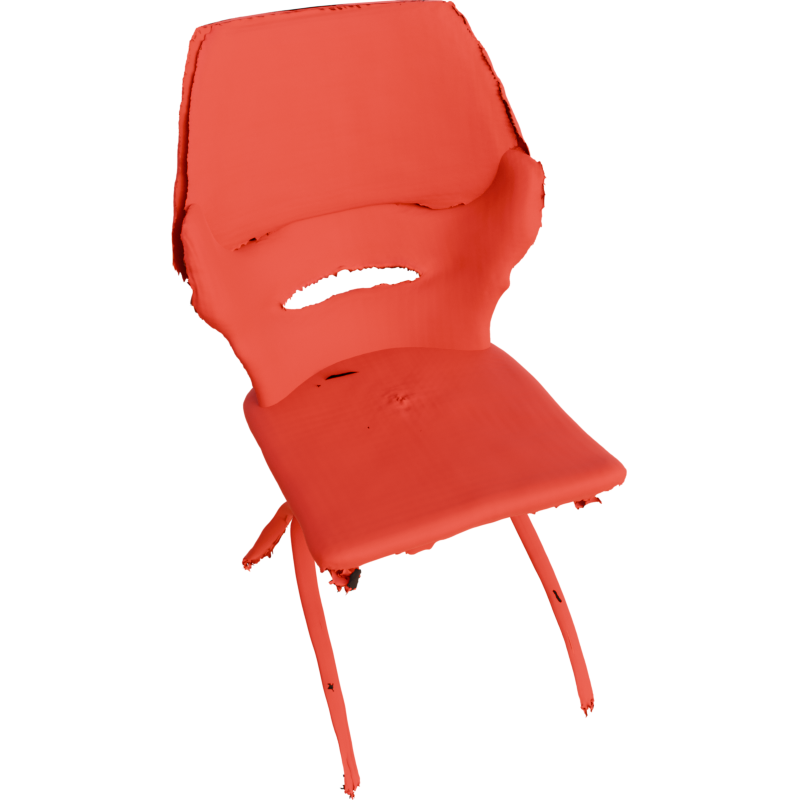}
    \includegraphics[width=\linewidth,height=\linewidth]{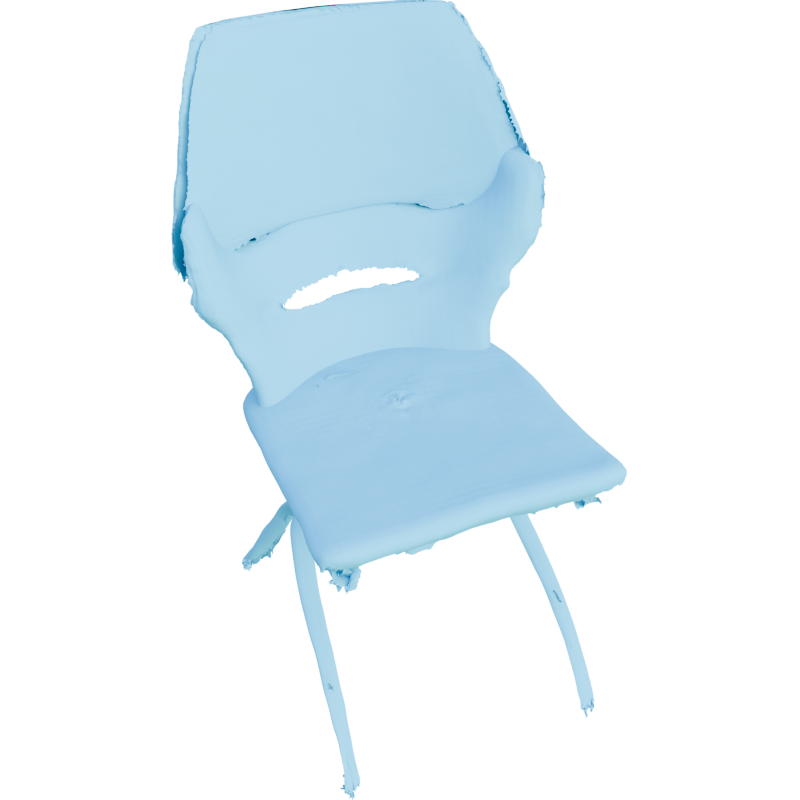}\vspace{1mm}
    \parbox{\linewidth}{\centering\small\textbf{Seed:} \textit{454712}}
\end{minipage}\hfill
\begin{minipage}[t]{0.32\linewidth}
    \centering
    \includegraphics[width=\linewidth,height=\linewidth]{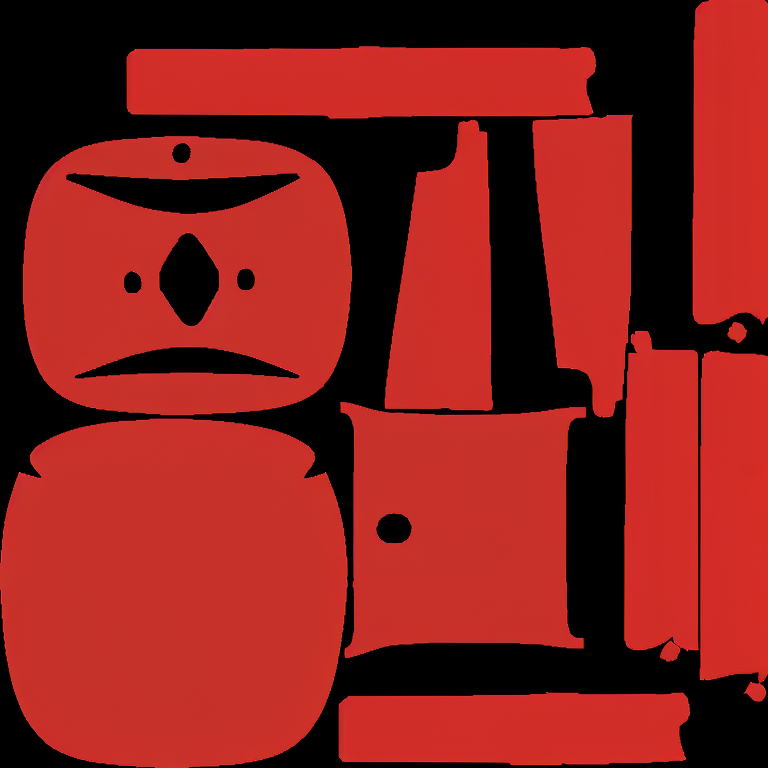}
    \includegraphics[width=\linewidth,height=\linewidth]{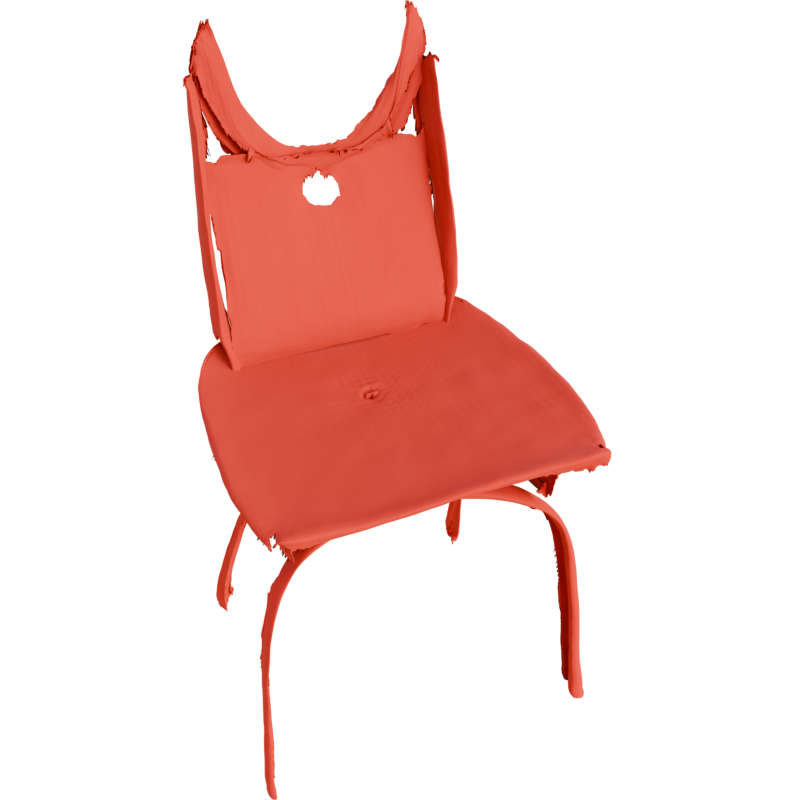}
    \includegraphics[width=\linewidth,height=\linewidth]{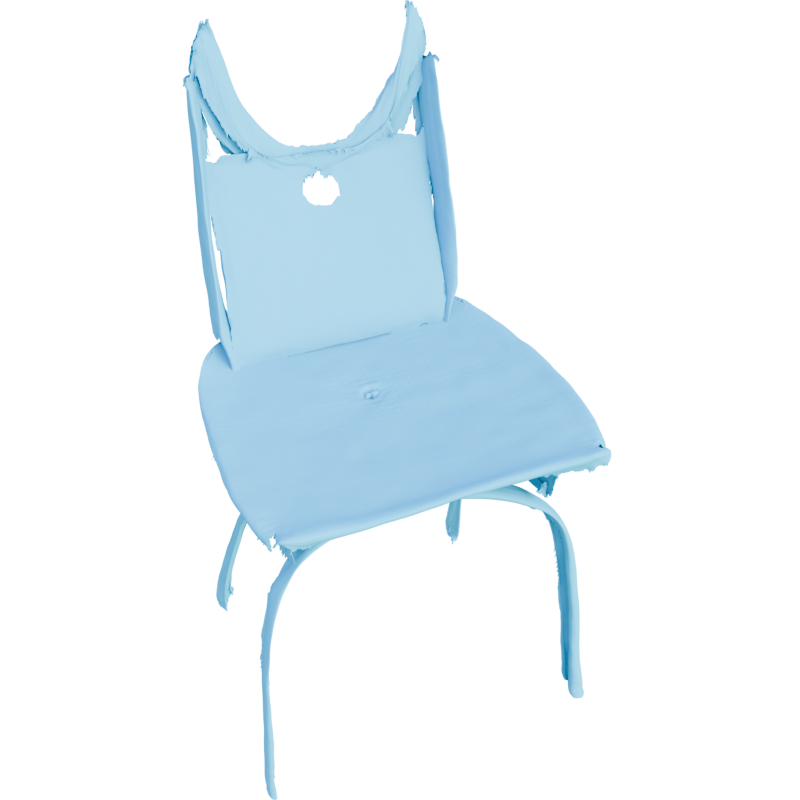}\vspace{1mm}
    \parbox{\linewidth}{\centering\small\textbf{Seed:} \textit{352231}}
\end{minipage}\hfill
\begin{minipage}[t]{0.32\linewidth}
    \centering
    \includegraphics[width=\linewidth,height=\linewidth]{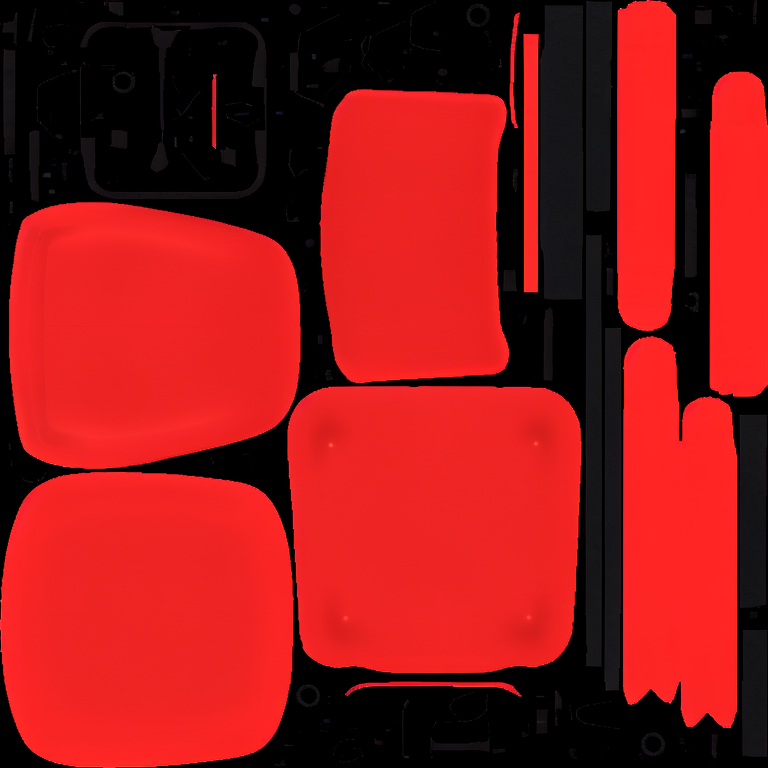}
    \includegraphics[width=\linewidth,height=\linewidth]{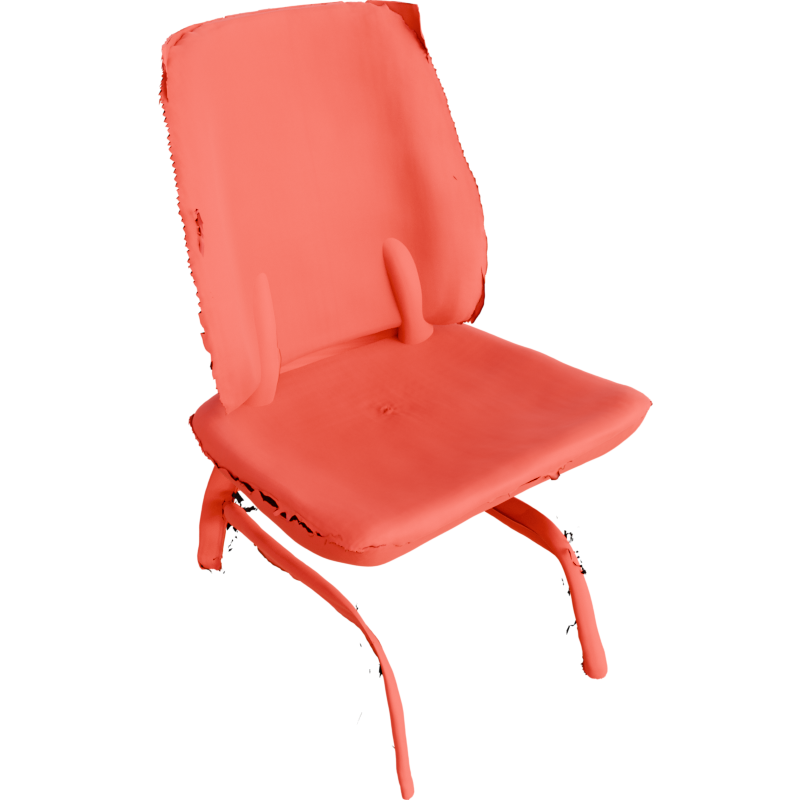}
    \includegraphics[width=\linewidth,height=\linewidth]{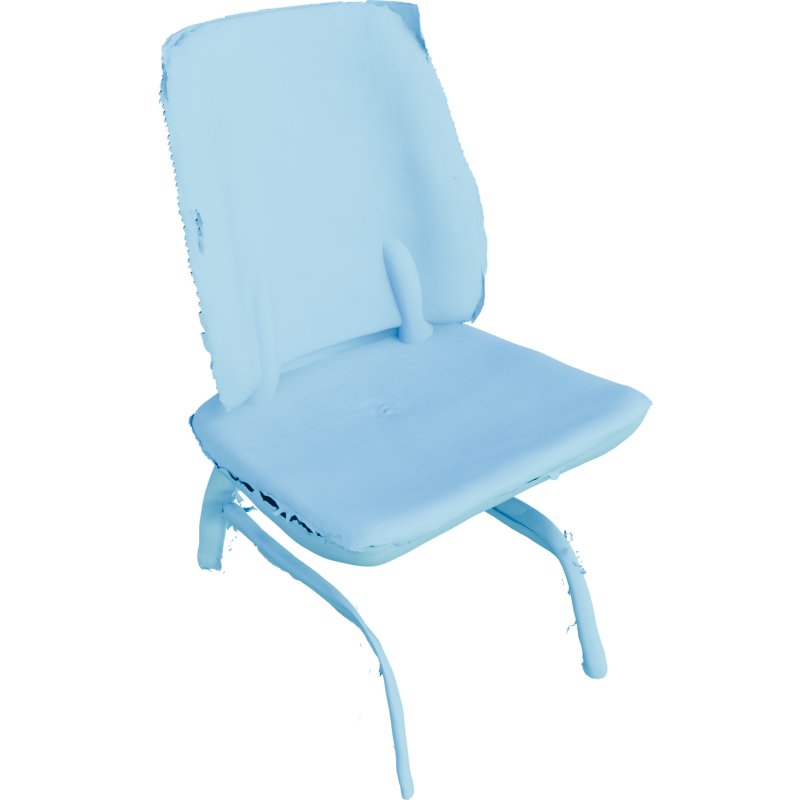}\vspace{1mm}
    \parbox{\linewidth}{\centering\small\textbf{Seed:} \textit{161715}}
\end{minipage}\vspace{2mm}
\begin{minipage}{0.9\linewidth}
    \centering\small\textbf{Prompt:} \textit{A red plastic chair}
\end{minipage}
\end{minipage}
\caption{Sample diversity of GIMDiffusion for minor changes to the prompt or different random seeds for the initial gaussian noise. It is clear that the generated variations differ significantly, not just in appearance and structure but also in the texture's atlas layout. This is invaluable in practical applications, where users typically generate multiple options and pick the best one.}\label{fig:gim-diversity}
\end{figure*}

\Cref{fig:gim-gen} shows the results of our method on a set of text prompts, generating objects as they might reasonably be queried, for example, in a gaming workflow.
The objects are \textbf{well-defined} and can be \textbf{relit from any direction}, as the generated albedo textures do not contain any lighting-related artifacts. Furthermore, \cref{fig:gim-diversity} illustrates that our proposed method is able to generate non-trivial variations for seed or prompt perturbations, which is crucial for the practical use of such a system.

\begin{figure}\centering
    \begin{minipage}[t]{0.48\linewidth}
        \centering
        \includegraphics[width=0.85\linewidth,height=0.85\linewidth]{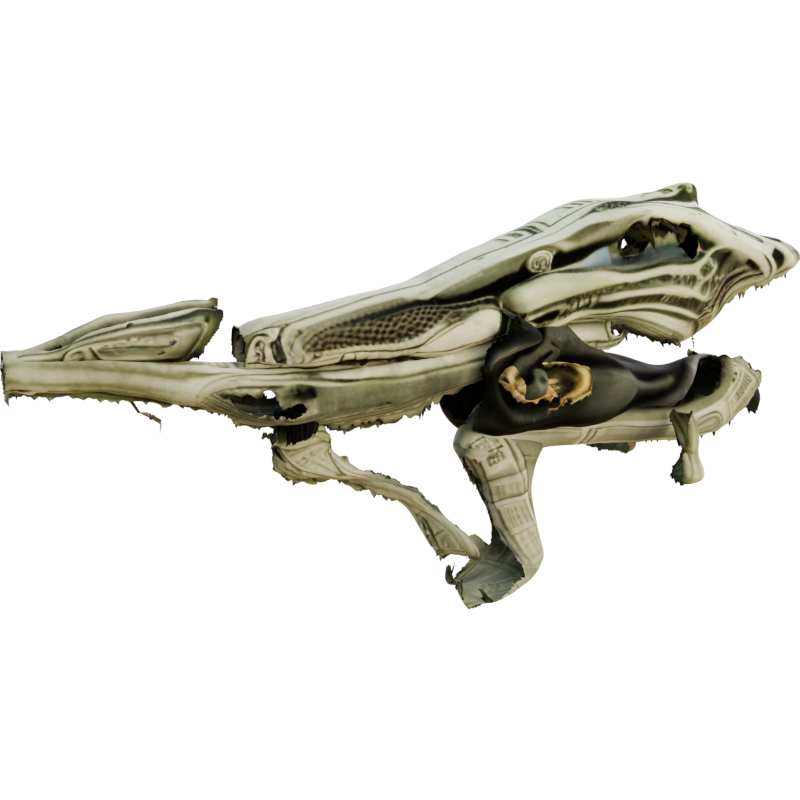}
        \parbox{0.95\linewidth}{\centering\footnotesize\textit{An organic alien gun is style of H.R. Giger}}
    \end{minipage}\hfill
    \begin{minipage}[t]{0.48\linewidth}
        \centering
        \includegraphics[width=0.85\linewidth,height=0.85\linewidth]{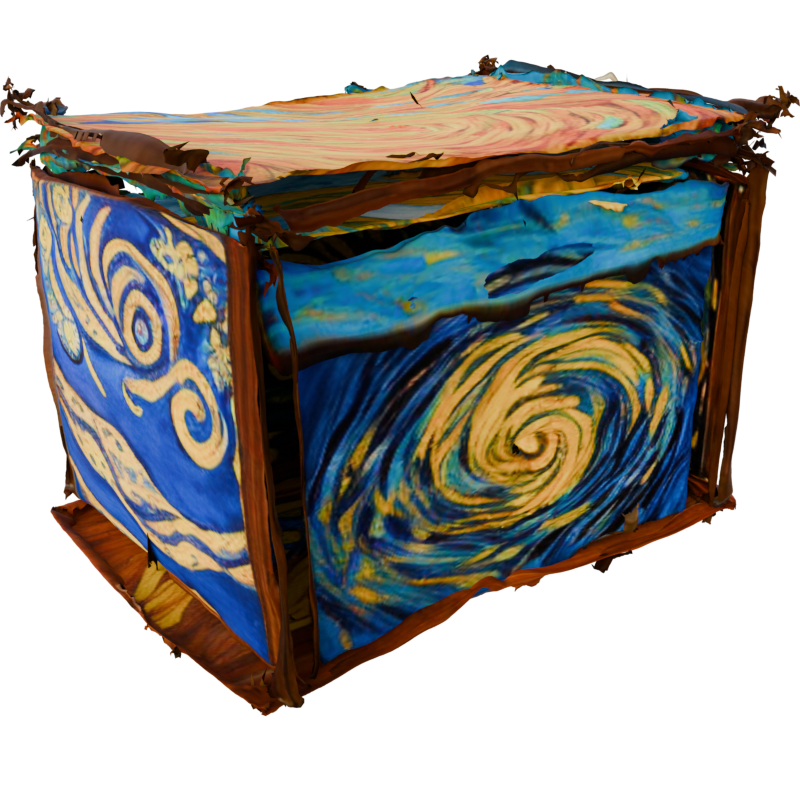}
        \parbox{0.95\linewidth}{\centering\footnotesize\textit{Wooden chest with Van Gogh's Starry Night painted on it}}
    \end{minipage}\vspace{1mm}
    \begin{minipage}[t]{0.48\linewidth}
        \centering
        \includegraphics[width=0.85\linewidth,height=0.85\linewidth]{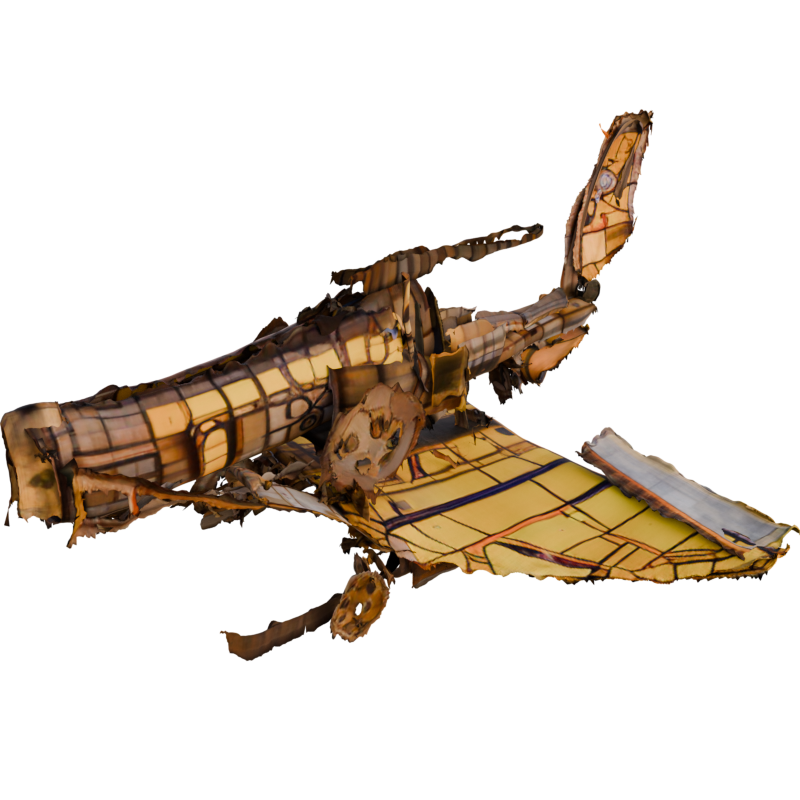}
        \parbox{0.95\linewidth}{\centering\footnotesize\textit{A steampunk airplane}}
    \end{minipage}\hfill
    \begin{minipage}[t]{0.48\linewidth}
        \centering
        \includegraphics[width=0.85\linewidth,height=0.85\linewidth]{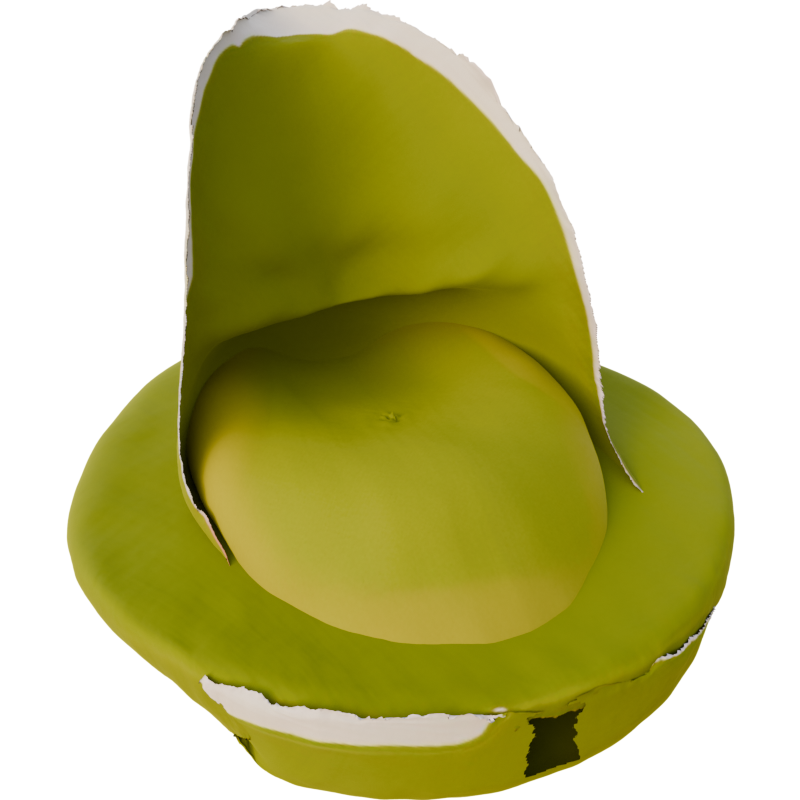}
        \parbox{0.95\linewidth}{\centering\footnotesize\textit{An avocado-shaped chair}}
    \end{minipage}
\caption{GIMDiffusion, thanks to the powerful natural image prior of the base model in combination with the Collaborative Control scheme, generalizes well outside the ``vanilla'' nature of the Objaverse training data.}\label{fig:gim-out-of-distribution}
\end{figure}

A key observation we make with regard to the concurrent work of Yan \etal{}~\cite{Yan2024AnOI} is that their method exhibits only very little variation in the UV layout of the generated objects within a given class. In contrast, GIMDiffusion shows significantly different UV layouts when either the seed or the prompt is varied slightly, which increases the practical value of generated multiple objects.
Finally, benefiting from the rich natural image prior, our model generalizes well beyond the initial 3D dataset.
Additional examples demonstrating this generalization are shown in \cref{fig:gim-out-of-distribution}.

\begin{figure}\centering
    \begin{minipage}[c]{0.48\linewidth}
        \centering
        \includegraphics[width=0.85\linewidth,height=0.85\linewidth]{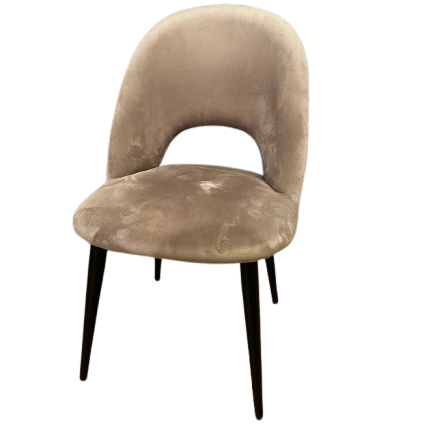}
        \parbox{0.95\linewidth}{\centering\footnotesize\phantom{\textit{Wooden chest with Van Gogh's}} \phantom{\textit{Starry Night painted on it}}}
    \end{minipage}\hfill
    \begin{minipage}[c]{0.48\linewidth}
        \centering
        \includegraphics[width=0.85\linewidth,height=0.85\linewidth]{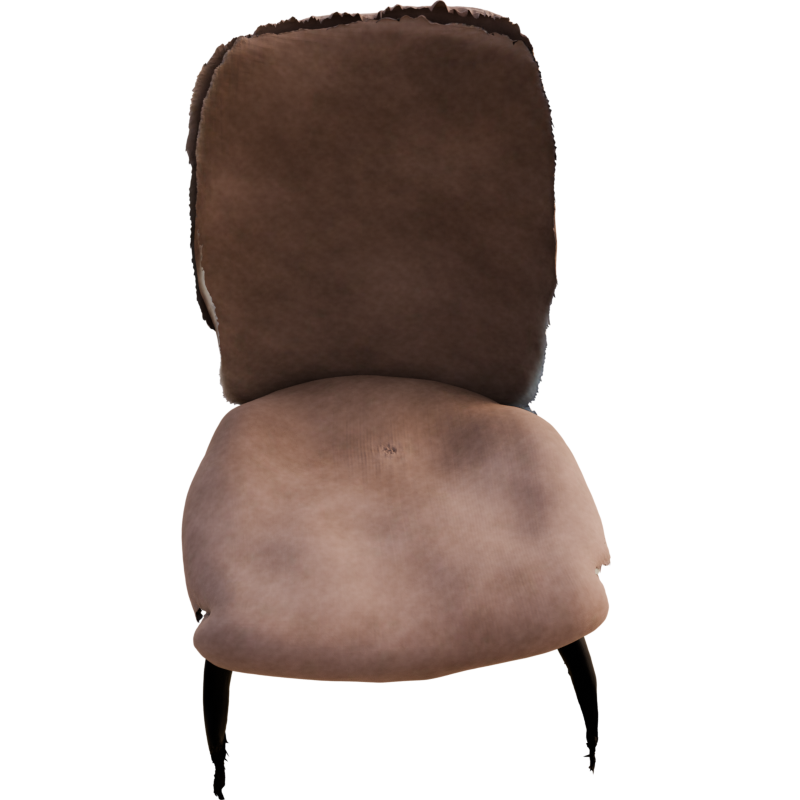}
        \parbox{0.95\linewidth}{\centering\footnotesize\textit{<No prompt>}}
    \end{minipage}\vspace{1mm}
    \begin{minipage}[t]{0.48\linewidth}
        \centering
        \includegraphics[width=0.85\linewidth,height=0.85\linewidth]{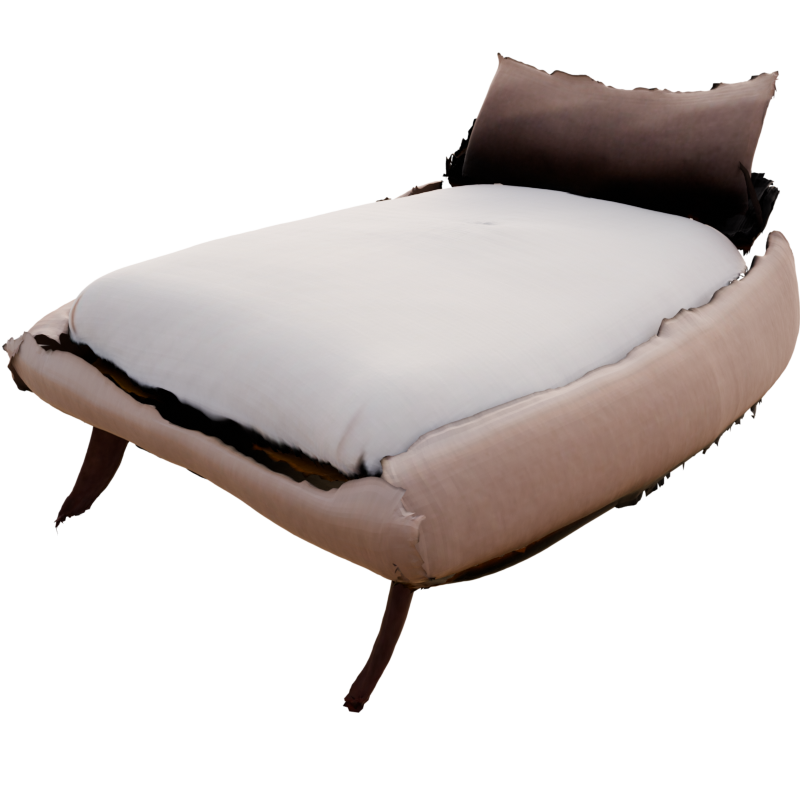}
        \parbox{0.95\linewidth}{\centering\footnotesize\textit{A kingsize bed.}}
    \end{minipage}\hfill
    \begin{minipage}[t]{0.48\linewidth}
        \centering
        \includegraphics[width=0.85\linewidth,height=0.85\linewidth]{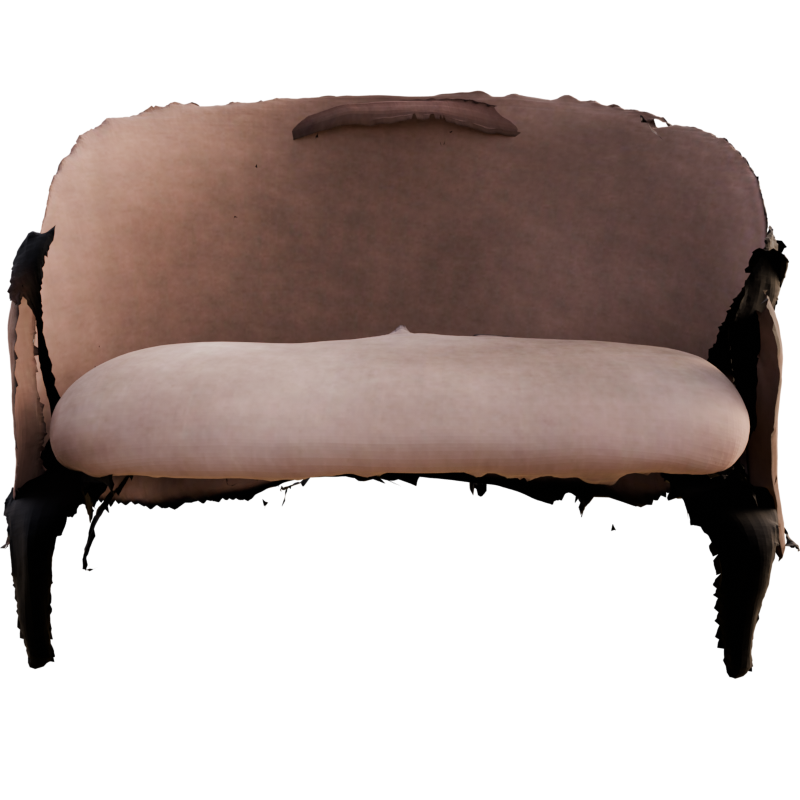}
        \parbox{0.95\linewidth}{\centering\footnotesize\textit{A sofa.}}
    \end{minipage}
\caption{We can stylistically guide the reverse process by applying a pre-trained IPAdapter to the frozen base model generation the albedo textures~\cite{ye2023ip-adapter}. This is extremely valuable in applications where the assets need to match an existing ``feel''.}\label{fig:ip-adapter-results}
\end{figure}

\begin{figure*}
\centering
\begin{minipage}[c]{0.32\linewidth}
    \centering
    \includegraphics[width=0.8\linewidth,height=0.8\linewidth]{figures/gim_diversity/mushroom_texture.png}
\end{minipage}\hfill
\begin{minipage}[c]{0.32\linewidth}
    \centering
    \includegraphics[width=\linewidth,height=\linewidth]{figures/gim_diversity/mushroom.png}
\end{minipage}\hfill
\begin{minipage}[c]{0.32\linewidth}
    \centering
    \includegraphics[width=\linewidth,height=\linewidth]{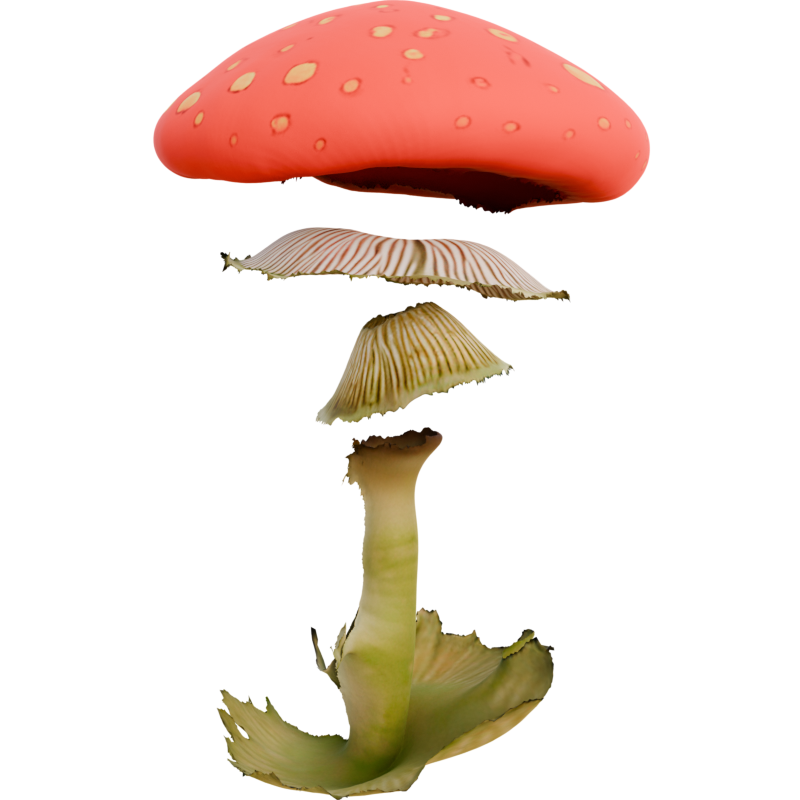}
\end{minipage}\vspace{2mm}
\begin{minipage}[c]{0.32\linewidth}\centering
    \parbox{\linewidth}{\centering (a) Albedo texture}
\end{minipage}\hfill
\begin{minipage}[c]{0.32\linewidth}\centering
    \parbox{\linewidth}{\centering (b) Textured mesh}
\end{minipage}\hfill
\begin{minipage}[c]{0.32\linewidth}\centering
    \parbox{\linewidth}{\centering (c) Exploded view of the geometry}
\end{minipage}
\caption{The generated images retain the semantically meaningful separation of charts in the texture atlas. We illustrate this here by showing an exploded view of a generated ``A fly
agaric mushroom''. It is clear that the various parts of the mushroom were separated as a human might, and that the method is even able to model internal parts of the shape.}\label{fig:gim-blow-up}
\end{figure*}

\subsection{IPAdapter Compatibility}

In practical terms, guiding the style of generated objects is crucial for many applications.
Major efforts have been made to achieve style control in diffusion models~\cite{ye2023ip-adapter}, and our method is compatible with these techniques, similar to the original Collaborative Control approach~\cite{Boss2024CollaborativeCF}.
As we are leveraging a frozen base model to generate the albedo textures, we can apply a pre-trained IPAdapter to that base model and produce stylized output meshes, as shown in \cref{fig:ip-adapter-results}.
Despite the significant mismatch in the application domain (natural images compared to albedo atlases), the style guidance remains successful.
We find that this approach starts to break down when the text prompt deviates too much (structurally) from the content of the conditioning image, which we attribute to the fact that IPAdapter aims to leverage every aspect of the conditioning image; but \eg{} IPAdapterInstruct~\cite{rowles2024ipadapterinstruct} offers a selective extraction of just appearance style but without entangling structure.

\subsection{Separability and Internal Structure}

A key advantage of our method is that it generates objects divided into distinct semantic (or nearly semantic) parts as shown in \cref{fig:gim-blow-up}, making the generated objects more suitable for editing and animation.
This capability arises from the multi-chart representation design and the semantic information embedded in the training data through handcrafted UV-maps, which loosely correspond to the semantic components of objects. This approach also allows users to easily correct imperfections, such as misaligned parts or extraneous geometry, and even combine different parts from multiple generations. 
Additionally, our method generates internal structures, such as the filament inside a light bulb or the interior of a fish tank, because geometry images represent the entire object holistically, not just the visible surfaces. 

\subsection{Limitations}\label{sec:limitations}

Our current method is not without limitations.
The most common issue is the appearance of visible cracks in the generated meshes.
While we do not currently stitch the charts' seams together, which could improve the visual quality of the generated meshes, we believe that this problem is further exacerbated by the VAE's latent compression.
Areas smaller than $8\times 8$ pixels are essentially below the VAE's latent resolution, causing visual problems.

We also encounter considerable ambiguity in the geometry image mapping.
Although individual charts in a geometry image can be rotated arbitrarily and still represent the same 3D object, the frozen base model is not rotationally equivariant~\cite{Weiler2021CoordinateIC}.
Human faces serve as a good example --- while typically upright in natural images, they can appear in random orientations on texture maps, which leads to varying (and overall lower) quality for such prompts.

Additionally, we find that the model sometimes duplicates parts of the object, which can lead to visual artifacts in the generated meshes due to z-fighting.
However, the separated nature of the generated objects makes these artifacts relatively easy to resolve manually.
\section{Discussion and future work}

In this work, we present \textbf{Geometry Image Diffusion }(GIMDiffusion), a novel Text-to-3D generation paradigm that utilizes geometry images as its core 3D representation in combination with powerful natural image priors in the form of pre-trained diffusion models.
Our results show that GIMDiffusion can generate relightable 3D assets as efficiently as existing Text-to-Image methods generate normal images, while avoiding the need for complex, custom 3D-aware architectures. We believe that our research lays the groundwork for a new direction in Text-to-3D generation.

Further quality improvements include addressing issues such as inter-chart alignment and eliminating visible cracks.
Additionally, incorporating topology prediction and conditioning on specific polygon budgets would enhance control over the generated 3D objects, making them more suitable for use in gaming and other graphics pipelines.
Equally promising is the potential of GIMDiffusion in related fields such as animation or text-to-video generation.

\section{Acknowledgements}

We would like to thank Shimon Vainer from Unity Technologies, Dr. Lev Melnikovsky from the Weizmann Institute and Alexander Demidko for their valuable feedback and insightful discussions. Konstantin Kutsy has been invaluable in helping us operate our training infrastructure.

{
    \small
    \bibliographystyle{ieeenat_fullname}
    \bibliography{library,add}

\begin{thebibliography}{69}
\providecommand{\natexlab}[1]{#1}
\providecommand{\url}[1]{\texttt{#1}}
\expandafter\ifx\csname urlstyle\endcsname\relax
  \providecommand{\doi}[1]{doi: #1}\else
  \providecommand{\doi}{doi: \begingroup \urlstyle{rm}\Url}\fi

\bibitem[Alhaija et~al.(2022)Alhaija, Dirik, Knorig, Fidler, and Shugrina]{Alhaija2022XDGANM3}
Hassan~Abu Alhaija, Alara Dirik, Andr'e Knorig, Sanja Fidler, and Maria Shugrina.
\newblock Xdgan: Multi-modal 3d shape generation in 2d space.
\newblock In \emph{British Machine Vision Conference}, 2022.

\bibitem[Alldieck et~al.(2024)Alldieck, Kolotouros, and Sminchisescu]{Alldieck2024ScoreDS}
Thiemo Alldieck, Nikos Kolotouros, and Cristian Sminchisescu.
\newblock Score distillation sampling with learned manifold corrective.
\newblock \emph{ArXiv}, abs/2401.05293, 2024.

\bibitem[Bensadoun et~al.(2024)Bensadoun, Monnier, Kleiman, Kokkinos, Siddiqui, Kariya, Harosh, Shapovalov, Graham, Garreau, Karnewar, Cao, Azuri, Makarov, Le, Toisoul, Novotny, Gafni, Neverova, and Vedaldi]{bensadoun2024meta3dgen}
Raphael Bensadoun, Tom Monnier, Yanir Kleiman, Filippos Kokkinos, Yawar Siddiqui, Mahendra Kariya, Omri Harosh, Roman Shapovalov, Benjamin Graham, Emilien Garreau, Animesh Karnewar, Ang Cao, Idan Azuri, Iurii Makarov, Eric-Tuan Le, Antoine Toisoul, David Novotny, Oran Gafni, Natalia Neverova, and Andrea Vedaldi.
\newblock Meta 3d gen, 2024.

\bibitem[Boss et~al.(2024)Boss, Huang, Vasishta, and Jampani]{Boss2024SF3DSF}
Mark Boss, Zixuan Huang, Aaryaman Vasishta, and Varun Jampani.
\newblock Sf3d: Stable fast 3d mesh reconstruction with uv-unwrapping and illumination disentanglement.
\newblock 2024.

\bibitem[Chan et~al.(2021)Chan, Lin, Chan, Nagano, Pan, Mello, Gallo, Guibas, Tremblay, Khamis, Karras, and Wetzstein]{Chan2021EfficientG3}
Eric Chan, Connor~Z. Lin, Matthew Chan, Koki Nagano, Boxiao Pan, Shalini~De Mello, Orazio Gallo, Leonidas~J. Guibas, Jonathan Tremblay, S. Khamis, Tero Karras, and Gordon Wetzstein.
\newblock Efficient geometry-aware 3d generative adversarial networks.
\newblock \emph{2022 IEEE/CVF Conference on Computer Vision and Pattern Recognition (CVPR)}, pages 16102--16112, 2021.

\bibitem[Chen and Zhang(2018)]{Chen2018LearningIF}
Zhiqin Chen and Hao Zhang.
\newblock Learning implicit fields for generative shape modeling.
\newblock \emph{2019 IEEE/CVF Conference on Computer Vision and Pattern Recognition (CVPR)}, pages 5932--5941, 2018.

\bibitem[Collins et~al.(2021)Collins, Goel, Luthra, Xu, Deng, Zhang, Vicente, Arora, Dideriksen, Guillaumin, and Malik]{Collins2021ABODA}
Jasmine Collins, Shubham Goel, Achleshwar Luthra, Leon~L. Xu, Kenan Deng, Xi Zhang, T.~F.~Y. Vicente, Himanshu Arora, T.~L. Dideriksen, Matthieu Guillaumin, and Jitendra Malik.
\newblock Abo: Dataset and benchmarks for real-world 3d object understanding.
\newblock \emph{2022 IEEE/CVF Conference on Computer Vision and Pattern Recognition (CVPR)}, pages 21094--21104, 2021.

\bibitem[Deitke et~al.(2022)Deitke, Schwenk, Salvador, Weihs, Michel, VanderBilt, Schmidt, Ehsani, Kembhavi, and Farhadi]{Deitke2022ObjaverseAU}
Matt Deitke, Dustin Schwenk, Jordi Salvador, Luca Weihs, Oscar Michel, Eli VanderBilt, Ludwig Schmidt, Kiana Ehsani, Aniruddha Kembhavi, and Ali Farhadi.
\newblock Objaverse: A universe of annotated 3d objects.
\newblock \emph{2023 IEEE/CVF Conference on Computer Vision and Pattern Recognition (CVPR)}, pages 13142--13153, 2022.

\bibitem[Doi and Koide(1991)]{Doi1991AnEM}
Akio Doi and Akio Koide.
\newblock An efficient method of triangulating equi-valued surfaces by using tetrahedral cells.
\newblock \emph{IEICE Transactions on Information and Systems}, 74:\penalty0 214--224, 1991.

\bibitem[Duan et~al.(2023)Duan, Guo, and Zhu]{Duan2023DiffusionDepthDD}
Yiqun Duan, Xianda Guo, and Zhengbiao Zhu.
\newblock Diffusiondepth: Diffusion denoising approach for monocular depth estimation.
\newblock \emph{ArXiv}, abs/2303.05021, 2023.

\bibitem[Esser et~al.(2024)Esser, Kulal, Blattmann, Entezari, Muller, Saini, Levi, Lorenz, Sauer, Boesel, Podell, Dockhorn, English, Lacey, Goodwin, Marek, and Rombach]{Esser2024ScalingRF}
Patrick Esser, Sumith Kulal, A. Blattmann, Rahim Entezari, Jonas Muller, Harry Saini, Yam Levi, Dominik Lorenz, Axel Sauer, Frederic Boesel, Dustin Podell, Tim Dockhorn, Zion English, Kyle Lacey, Alex Goodwin, Yannik Marek, and Robin Rombach.
\newblock Scaling rectified flow transformers for high-resolution image synthesis.
\newblock \emph{ArXiv}, abs/2403.03206, 2024.

\bibitem[Gu et~al.(2002)Gu, Gortler, and Hoppe]{Gu2002GeometryI}
Xianfeng Gu, Steven~J. Gortler, and Hugues Hoppe.
\newblock Geometry images.
\newblock \emph{Proceedings of the 29th annual conference on Computer graphics and interactive techniques}, 2002.

\bibitem[H{\"o}llein et~al.(2024)H{\"o}llein, Bovzivc, Muller, Novotny, Tseng, Richardt, Zollhofer, and Nie{\ss}ner]{Hllein2024ViewDiff3I}
Lukas H{\"o}llein, Aljavz Bovzivc, Norman Muller, David Novotny, Hung-Yu Tseng, Christian Richardt, Michael Zollhofer, and Matthias Nie{\ss}ner.
\newblock Viewdiff: 3d-consistent image generation with text-to-image models.
\newblock \emph{ArXiv}, abs/2403.01807, 2024.

\bibitem[Hong et~al.(2024)Hong, Tang, Cao, Shi, Wu, Chen, Wang, Pan, Lin, and Liu]{Hong20243DTopiaLT}
Fangzhou Hong, Jiaxiang Tang, Ziang Cao, Min Shi, Tong Wu, Zhaoxi Chen, Tengfei Wang, Liang Pan, Dahua Lin, and Ziwei Liu.
\newblock 3dtopia: Large text-to-3d generation model with hybrid diffusion priors.
\newblock \emph{ArXiv}, abs/2403.02234, 2024.

\bibitem[Hong et~al.(2023)Hong, Zhang, Gu, Bi, Zhou, Liu, Liu, Sunkavalli, Bui, and Tan]{Hong2023LRMLR}
Yicong Hong, Kai Zhang, Jiuxiang Gu, Sai Bi, Yang Zhou, Difan Liu, Feng Liu, Kalyan Sunkavalli, Trung Bui, and Hao Tan.
\newblock Lrm: Large reconstruction model for single image to 3d.
\newblock \emph{ArXiv}, abs/2311.04400, 2023.

\bibitem[Hu et~al.(2023)Hu, Gao, Zhang, Sun, Zhang, and Bo]{Hu2023AnimateAC}
Liucheng Hu, Xin Gao, Peng Zhang, Ke Sun, Bang Zhang, and Liefeng Bo.
\newblock Animate anyone: Consistent and controllable image-to-video synthesis for character animation.
\newblock \emph{ArXiv}, abs/2311.17117, 2023.

\bibitem[Huang et~al.(2024)Huang, Johnson, Debnath, Rehg, and Wu]{Huang2024PointInfinityRP}
Zixuan Huang, Justin Johnson, Shoubhik Debnath, James~M. Rehg, and Chao-Yuan Wu.
\newblock Pointinfinity: Resolution-invariant point diffusion models.
\newblock 2024.

\bibitem[Jun and Nichol(2023)]{Jun2023ShapEGC}
Heewoo Jun and Alex Nichol.
\newblock Shap-e: Generating conditional 3d implicit functions.
\newblock \emph{ArXiv}, abs/2305.02463, 2023.

\bibitem[Kant et~al.(2024)Kant, Wu, Vasilkovsky, Qian, Ren, Guler, Ghanem, Tulyakov, Gilitschenski, and Siarohin]{Kant2024SPADS}
Yash Kant, Ziyi Wu, Michael Vasilkovsky, Guocheng Qian, Jian Ren, Riza~Alp Guler, Bernard Ghanem, S. Tulyakov, Igor Gilitschenski, and Aliaksandr Siarohin.
\newblock Spad : Spatially aware multiview diffusers.
\newblock \emph{ArXiv}, abs/2402.05235, 2024.

\bibitem[Karras et~al.(2018)Karras, Laine, and Aila]{Karras2018ASG}
Tero Karras, Samuli Laine, and Timo Aila.
\newblock A style-based generator architecture for generative adversarial networks.
\newblock \emph{2019 IEEE/CVF Conference on Computer Vision and Pattern Recognition (CVPR)}, pages 4396--4405, 2018.

\bibitem[Katzir et~al.(2023)Katzir, Patashnik, Cohen-Or, and Lischinski]{Katzir2023NoiseFreeSD}
Oren Katzir, Or Patashnik, Daniel Cohen-Or, and Dani Lischinski.
\newblock Noise-free score distillation.
\newblock \emph{ArXiv}, abs/2310.17590, 2023.

\bibitem[Ke et~al.(2023)Ke, Obukhov, Huang, Metzger, Daudt, and Schindler]{Ke2023RepurposingDI}
Bing~Wen Ke, Anton Obukhov, Shengyu Huang, Nando Metzger, Rodrigo~Caye Daudt, and Konrad Schindler.
\newblock Repurposing diffusion-based image generators for monocular depth estimation.
\newblock \emph{ArXiv}, abs/2312.02145, 2023.

\bibitem[Kerbl et~al.(2023)Kerbl, Kopanas, Leimkuehler, and Drettakis]{Kerbl20233DGS}
Bernhard Kerbl, Georgios Kopanas, Thomas Leimkuehler, and George Drettakis.
\newblock 3d gaussian splatting for real-time radiance field rendering.
\newblock \emph{ACM Transactions on Graphics (TOG)}, 42:\penalty0 1 -- 14, 2023.

\bibitem[Liang et~al.(2023)Liang, Yang, Lin, Li, Xu, and Chen]{Liang2023LucidDreamerTH}
Yixun Liang, Xin Yang, Jiantao Lin, Haodong Li, Xiaogang Xu, and Yingcong Chen.
\newblock Luciddreamer: Towards high-fidelity text-to-3d generation via interval score matching.
\newblock \emph{ArXiv}, abs/2311.11284, 2023.

\bibitem[Lin et~al.(2024)Lin, Liu, Li, and Yang]{lin2024zerosnr}
Shanchuan Lin, Bingchen Liu, Jiashi Li, and Xiao Yang.
\newblock Common diffusion noise schedules and sample steps are flawed.
\newblock In \emph{Proceedings of the IEEE/CVF Winter Conference on Applications of Computer Vision}, pages 5404--5411, 2024.

\bibitem[Lipman et~al.(2022)Lipman, Chen, Ben-Hamu, Nickel, and Le]{Lipman2022-wr}
Yaron Lipman, Ricky T~Q Chen, Heli Ben-Hamu, Maximilian Nickel, and Matt Le.
\newblock Flow matching for generative modeling.
\newblock 2022.

\bibitem[Liu et~al.(2023)Liu, Wu, Hoorick, Tokmakov, Zakharov, and Vondrick]{Liu2023Zero1to3ZO}
Ruoshi Liu, Rundi Wu, Basile~Van Hoorick, Pavel Tokmakov, Sergey Zakharov, and Carl Vondrick.
\newblock Zero-1-to-3: Zero-shot one image to 3d object.
\newblock \emph{2023 IEEE/CVF International Conference on Computer Vision (ICCV)}, pages 9264--9275, 2023.

\bibitem[Lorensen and Cline(1987)]{Lorensen1987MarchingCA}
William~E. Lorensen and Harvey~E. Cline.
\newblock Marching cubes: A high resolution 3d surface construction algorithm.
\newblock \emph{Proceedings of the 14th annual conference on Computer graphics and interactive techniques}, 1987.

\bibitem[Lorraine et~al.(2023)Lorraine, Xie, Zeng, Lin, Takikawa, Sharp, Lin, Liu, Fidler, and Lucas]{Lorraine2023ATT3DAT}
Jonathan Lorraine, Kevin Xie, Xiaohui Zeng, Chen-Hsuan Lin, Towaki Takikawa, Nicholas Sharp, Tsung-Yi Lin, Ming-Yu Liu, Sanja Fidler, and James Lucas.
\newblock Att3d: Amortized text-to-3d object synthesis.
\newblock \emph{2023 IEEE/CVF International Conference on Computer Vision (ICCV)}, pages 17900--17910, 2023.

\bibitem[Luo et~al.(2023)Luo, Rockwell, Lee, and Johnson]{luo2023scalable}
Tiange Luo, Chris Rockwell, Honglak Lee, and Justin Johnson.
\newblock Scalable 3d captioning with pretrained models.
\newblock \emph{arXiv preprint arXiv:2306.07279}, 2023.

\bibitem[Malladi et~al.(1995)Malladi, Sethian, and Vemuri]{Malladi1995ShapeMW}
Ravi Malladi, James~A. Sethian, and Baba~C. Vemuri.
\newblock Shape modeling with front propagation: A level set approach.
\newblock \emph{IEEE Trans. Pattern Anal. Mach. Intell.}, 17:\penalty0 158--175, 1995.

\bibitem[Mescheder et~al.(2018)Mescheder, Oechsle, Niemeyer, Nowozin, and Geiger]{Mescheder2018OccupancyNL}
Lars~M. Mescheder, Michael Oechsle, Michael Niemeyer, Sebastian Nowozin, and Andreas Geiger.
\newblock Occupancy networks: Learning 3d reconstruction in function space.
\newblock \emph{2019 IEEE/CVF Conference on Computer Vision and Pattern Recognition (CVPR)}, pages 4455--4465, 2018.

\bibitem[Mildenhall et~al.(2020)Mildenhall, Srinivasan, Tancik, Barron, Ramamoorthi, and Ng]{Mildenhall2020NeRF}
Ben Mildenhall, Pratul~P. Srinivasan, Matthew Tancik, Jonathan~T. Barron, Ravi Ramamoorthi, and Ren Ng.
\newblock Nerf.
\newblock \emph{Communications of the ACM}, 65:\penalty0 99 -- 106, 2020.

\bibitem[Nichol et~al.(2022)Nichol, Jun, Dhariwal, Mishkin, and Chen]{Nichol2022PointEAS}
Alex Nichol, Heewoo Jun, Prafulla Dhariwal, Pamela Mishkin, and Mark Chen.
\newblock Point-e: A system for generating 3d point clouds from complex prompts.
\newblock \emph{ArXiv}, abs/2212.08751, 2022.

\bibitem[Pernias et~al.(2024)Pernias, Rampas, Richter, Pal, and Aubreville]{Pernias2024WrstchenAE}
Pablo Pernias, Dominic Rampas, Mats~L. Richter, Christopher Pal, and Marc Aubreville.
\newblock W{\"u}rstchen: An efficient architecture for large-scale text-to-image diffusion models.
\newblock In \emph{International Conference on Learning Representations}, 2024.

\bibitem[Podell et~al.(2023)Podell, English, Lacey, Blattmann, Dockhorn, Muller, Penna, and Rombach]{Podell2023SDXLIL}
Dustin Podell, Zion English, Kyle Lacey, A. Blattmann, Tim Dockhorn, Jonas Muller, Joe Penna, and Robin Rombach.
\newblock Sdxl: Improving latent diffusion models for high-resolution image synthesis.
\newblock \emph{ArXiv}, abs/2307.01952, 2023.

\bibitem[Poole et~al.(2022)Poole, Jain, Barron, and Mildenhall]{Poole2022DreamFusionTU}
Ben Poole, Ajay Jain, Jonathan~T. Barron, and Ben Mildenhall.
\newblock Dreamfusion: Text-to-3d using 2d diffusion.
\newblock \emph{ArXiv}, abs/2209.14988, 2022.

\bibitem[Rombach et~al.(2021)Rombach, Blattmann, Lorenz, Esser, and Ommer]{Rombach2021HighResolutionIS}
Robin Rombach, A. Blattmann, Dominik Lorenz, Patrick Esser, and Bj{\"o}rn Ommer.
\newblock High-resolution image synthesis with latent diffusion models.
\newblock \emph{2022 IEEE/CVF Conference on Computer Vision and Pattern Recognition (CVPR)}, pages 10674--10685, 2021.

\bibitem[Rowles et~al.(2024)Rowles, Vainer, Nigris, Elizarov, Kutsy, and Donné]{rowles2024ipadapterinstruct}
Ciara Rowles, Shimon Vainer, Dante~De Nigris, Slava Elizarov, Konstantin Kutsy, and Simon Donné.
\newblock Ipadapter-instruct: Resolving ambiguity in image-based conditioning using instruct prompts, 2024.

\bibitem[Sander et~al.(2003)Sander, Wood, Gortler, Snyder, and Hoppe]{Sander2003MultiChartGI}
Pedro~V. Sander, Zo{\"e}~J. Wood, Steven~J. Gortler, John~M. Snyder, and Hugues Hoppe.
\newblock Multi-chart geometry images.
\newblock In \emph{Eurographics Symposium on Geometry Processing}, 2003.

\bibitem[Sawhney and Crane(2017)]{Sawhney2017BoundaryFF}
Rohan Sawhney and Keenan Crane.
\newblock Boundary first flattening.
\newblock \emph{ACM Transactions on Graphics (TOG)}, 37:\penalty0 1 -- 14, 2017.

\bibitem[Schuhmann et~al.(2022)Schuhmann, Beaumont, Vencu, Gordon, Wightman, Cherti, Coombes, Katta, Mullis, Wortsman, Schramowski, Kundurthy, Crowson, Schmidt, Kaczmarczyk, and Jitsev]{Schuhmann2022LAION5BAO}
Christoph Schuhmann, Romain Beaumont, Richard Vencu, Cade Gordon, Ross Wightman, Mehdi Cherti, Theo Coombes, Aarush Katta, Clayton Mullis, Mitchell Wortsman, Patrick Schramowski, Srivatsa Kundurthy, Katherine Crowson, Ludwig Schmidt, Robert Kaczmarczyk, and Jenia Jitsev.
\newblock Laion-5b: An open large-scale dataset for training next generation image-text models.
\newblock \emph{ArXiv}, abs/2210.08402, 2022.

\bibitem[Shi et~al.(2023{\natexlab{a}})Shi, Chen, Zhang, Liu, Xu, Wei, Chen, Zeng, and Su]{Shi2023Zero123AS}
Ruoxi Shi, Hansheng Chen, Zhuoyang Zhang, Minghua Liu, Chao Xu, Xinyue Wei, Linghao Chen, Chong Zeng, and Hao Su.
\newblock Zero123++: a single image to consistent multi-view diffusion base model.
\newblock \emph{ArXiv}, abs/2310.15110, 2023{\natexlab{a}}.

\bibitem[Shi et~al.(2023{\natexlab{b}})Shi, Wang, Ye, Long, Li, and Yang]{Shi2023MVDreamMD}
Yichun Shi, Peng Wang, Jianglong Ye, Mai Long, Kejie Li, and X. Yang.
\newblock Mvdream: Multi-view diffusion for 3d generation.
\newblock \emph{ArXiv}, abs/2308.16512, 2023{\natexlab{b}}.

\bibitem[Siddiqui et~al.(2023)Siddiqui, Alliegro, Artemov, Tommasi, Sirigatti, Rosov, Dai, and Nießner]{siddiqui2023meshgpt}
Yawar Siddiqui, Antonio Alliegro, Alexey Artemov, Tatiana Tommasi, Daniele Sirigatti, Vladislav Rosov, Angela Dai, and Matthias Nießner.
\newblock Meshgpt: Generating triangle meshes with decoder-only transformers, 2023.

\bibitem[Siddiqui et~al.(2024)Siddiqui, Monnier, Kokkinos, Kariya, Kleiman, Garreau, Gafni, Neverova, Vedaldi, Shapovalov, and Novotny]{Siddiqui2024Meta3A}
Yawar Siddiqui, Tom Monnier, Filippos Kokkinos, Mahendra Kariya, Yanir Kleiman, Emilien Garreau, Oran Gafni, Natalia~V. Neverova, Andrea Vedaldi, Roman Shapovalov, and David Novotny.
\newblock Meta 3d assetgen: Text-to-mesh generation with high-quality geometry, texture, and pbr materials.
\newblock \emph{ArXiv}, abs/2407.02445, 2024.

\bibitem[Sinha et~al.(2016)Sinha, Bai, and Ramani]{Sinha2016DeepL3}
Ayan Sinha, Jing Bai, and Karthik Ramani.
\newblock Deep learning 3d shape surfaces using geometry images.
\newblock In \emph{European Conference on Computer Vision}, 2016.

\bibitem[Sohl-Dickstein et~al.(2015)Sohl-Dickstein, Weiss, Maheswaranathan, and Ganguli]{Sohl-Dickstein2015-vg}
Jascha Sohl-Dickstein, Eric~A Weiss, Niru Maheswaranathan, and Surya Ganguli.
\newblock Deep unsupervised learning using nonequilibrium thermodynamics.
\newblock 2015.

\bibitem[Song et~al.(2020)Song, Sohl-Dickstein, Kingma, Kumar, Ermon, and Poole]{Song2020ScoreBasedGM}
Yang Song, Jascha~Narain Sohl-Dickstein, Diederik~P. Kingma, Abhishek Kumar, Stefano Ermon, and Ben Poole.
\newblock Score-based generative modeling through stochastic differential equations.
\newblock \emph{ArXiv}, abs/2011.13456, 2020.

\bibitem[Srinivasan et~al.(2023)Srinivasan, Garbin, Verbin, Barron, and Mildenhall]{Srinivasan2023NuvoNU}
Pratul~P. Srinivasan, Stephan~J. Garbin, Dor Verbin, Jonathan~T. Barron, and Ben Mildenhall.
\newblock Nuvo: Neural uv mapping for unruly 3d representations.
\newblock \emph{ArXiv}, abs/2312.05283, 2023.

\bibitem[Tochilkin et~al.(2024)Tochilkin, Pankratz, Liu, Huang, Letts, Li, Liang, Laforte, Jampani, and Cao]{Tochilkin2024TripoSRF3}
Dmitry Tochilkin, David Pankratz, Zexiang Liu, Zixuan Huang, Adam Letts, Yangguang Li, Ding Liang, Christian Laforte, Varun Jampani, and Yan-Pei Cao.
\newblock Triposr: Fast 3d object reconstruction from a single image.
\newblock \emph{ArXiv}, abs/2403.02151, 2024.

\bibitem[\url{https://huggingface.co/drhead}()]{drhead2024ZeroDiffusion}
\url{https://huggingface.co/drhead}.
\newblock Huggingface zerodiffusion model weights v0.9.
\newblock \url{https://huggingface.co/drhead/ZeroDiffusion}.
\newblock Accessed: 2024-02-08.

\bibitem[Vainer et~al.(2024)Vainer, Boss, Parger, Kutsy, Nigris, Rowles, Perony, and Donn'e]{Boss2024CollaborativeCF}
Shimon Vainer, Mark Boss, Mathias Parger, Konstantin Kutsy, Dante~De Nigris, Ciara Rowles, Nicolas Perony, and Simon Donn'e.
\newblock Collaborative control for geometry-conditioned pbr image generation.
\newblock \emph{ArXiv}, abs/2402.05919, 2024.

\bibitem[Vaswani et~al.(2017)Vaswani, Shazeer, Parmar, Uszkoreit, Jones, Gomez, Kaiser, and Polosukhin]{Vaswani2017AttentionIA}
Ashish Vaswani, Noam~M. Shazeer, Niki Parmar, Jakob Uszkoreit, Llion Jones, Aidan~N. Gomez, Lukasz Kaiser, and Illia Polosukhin.
\newblock Attention is all you need.
\newblock In \emph{Neural Information Processing Systems}, 2017.

\bibitem[Wang et~al.(2022)Wang, Du, Li, Yeh, and Shakhnarovich]{Wang2022ScoreJC}
Haochen Wang, Xiaodan Du, Jiahao Li, Raymond~A. Yeh, and Gregory Shakhnarovich.
\newblock Score jacobian chaining: Lifting pretrained 2d diffusion models for 3d generation.
\newblock \emph{2023 IEEE/CVF Conference on Computer Vision and Pattern Recognition (CVPR)}, pages 12619--12629, 2022.

\bibitem[Wang et~al.(2023)Wang, Lu, Wang, Bao, Li, Su, and Zhu]{Wang2023ProlificDreamerHA}
Zhengyi Wang, Cheng Lu, Yikai Wang, Fan Bao, Chongxuan Li, Hang Su, and Jun Zhu.
\newblock Prolificdreamer: High-fidelity and diverse text-to-3d generation with variational score distillation.
\newblock \emph{ArXiv}, abs/2305.16213, 2023.

\bibitem[Weiler et~al.(2021)Weiler, Forr'e, Verlinde, and Welling]{Weiler2021CoordinateIC}
Maurice Weiler, Patrick Forr'e, Erik~P. Verlinde, and Max Welling.
\newblock Coordinate independent convolutional networks - isometry and gauge equivariant convolutions on riemannian manifolds.
\newblock \emph{ArXiv}, abs/2106.06020, 2021.

\bibitem[Wu et~al.(2024)Wu, Zhou, Yi, Yuan, and Zhang]{Wu2024Consistent3DTC}
Zike Wu, Pan Zhou, Xuanyu Yi, Xiaoding Yuan, and Hanwang Zhang.
\newblock Consistent3d: Towards consistent high-fidelity text-to-3d generation with deterministic sampling prior.
\newblock \emph{ArXiv}, abs/2401.09050, 2024.

\bibitem[Xie et~al.(2024)Xie, Lorraine, Cao, Gao, Lucas, Torralba, Fidler, and Zeng]{Xie2024LATTE3DLA}
Kevin Xie, Jonathan Lorraine, Tianshi Cao, Jun Gao, James Lucas, Antonio Torralba, Sanja Fidler, and Xiaohui Zeng.
\newblock Latte3d: Large-scale amortized text-to-enhanced3d synthesis.
\newblock \emph{ArXiv}, abs/2403.15385, 2024.

\bibitem[Xie et~al.(2021)Xie, Takikawa, Saito, Litany, Yan, Khan, Tombari, Tompkin, Sitzmann, and Sridhar]{Xie2021NeuralFI}
Yiheng Xie, Towaki Takikawa, Shunsuke Saito, Or Litany, Shiqin Yan, Numair Khan, Federico Tombari, James Tompkin, Vincent Sitzmann, and Srinath Sridhar.
\newblock Neural fields in visual computing and beyond.
\newblock \emph{Computer Graphics Forum}, 41, 2021.

\bibitem[Yan et~al.(2024)Yan, Lee, Wan, and Chang]{Yan2024AnOI}
Xingguang Yan, Han-Hung Lee, Ziyu Wan, and Angel~X. Chang.
\newblock An object is worth 64x64 pixels: Generating 3d object via image diffusion.
\newblock 2024.

\bibitem[Ye et~al.(2023)Ye, Zhang, Liu, Han, and Yang]{ye2023ip-adapter}
Hu Ye, Jun Zhang, Sibo Liu, Xiao Han, and Wei Yang.
\newblock Ip-adapter: Text compatible image prompt adapter for text-to-image diffusion models.
\newblock \emph{arXiv preprint arXiv:2308.06721}, 2023.

\bibitem[Young(2022)]{Young2022}
Jonathan Young.
\newblock Xatlas: Mesh parameterization / uv unwrapping library.
\newblock \url{https://github.com/jpcy/xatlas}, 2022.

\bibitem[Zavadski et~al.(2023)Zavadski, Feiden, and Rother]{Zavadski2023ControlNetXSDA}
Denis Zavadski, Johann-Friedrich Feiden, and Carsten Rother.
\newblock Controlnet-xs: Designing an efficient and effective architecture for controlling text-to-image diffusion models.
\newblock \emph{ArXiv}, abs/2312.06573, 2023.

\bibitem[Zeng et~al.(2022)Zeng, Vahdat, Williams, Gojcic, Litany, Fidler, and Kreis]{Zeng2022LIONLP}
Xiaohui Zeng, Arash Vahdat, Francis Williams, Zan Gojcic, Or Litany, Sanja Fidler, and Karsten Kreis.
\newblock Lion: Latent point diffusion models for 3d shape generation.
\newblock \emph{ArXiv}, abs/2210.06978, 2022.

\bibitem[Zhang et~al.(2023)Zhang, Rao, and Agrawala]{Zhang2023AddingCC}
Lvmin Zhang, Anyi Rao, and Maneesh Agrawala.
\newblock Adding conditional control to text-to-image diffusion models.
\newblock \emph{2023 IEEE/CVF International Conference on Computer Vision (ICCV)}, pages 3813--3824, 2023.

\bibitem[Zheng and Vedaldi(2023)]{Zheng2023Free3DCN}
Chuanxia Zheng and Andrea Vedaldi.
\newblock Free3d: Consistent novel view synthesis without 3d representation.
\newblock \emph{ArXiv}, abs/2312.04551, 2023.

\bibitem[Zheng et~al.(2022)Zheng, Liu, Wang, and Tong]{Zheng2022SDFStyleGANIS}
Xin Zheng, Yang Liu, Peng-Shuai Wang, and Xin Tong.
\newblock Sdf‐stylegan: Implicit sdf‐based stylegan for 3d shape generation.
\newblock \emph{Computer Graphics Forum}, 41, 2022.

\bibitem[Zhu et~al.(2023)Zhu, Zhuang, and Koyejo]{Zhu2023HIFAHT}
Junzhe Zhu, Peiye Zhuang, and Oluwasanmi Koyejo.
\newblock Hifa: High-fidelity text-to-3d generation with advanced diffusion guidance.
\newblock In \emph{International Conference on Learning Representations}, 2023.

\end{thebibliography}
}

\end{document}